\documentclass[journal,twocolumn]{IEEEtran}
\usepackage{cite}
\usepackage{amsmath,amssymb,amsfonts,bm}

\usepackage{algorithmic}
\usepackage{graphicx}
\usepackage{subfigure}
\usepackage{textcomp}
\usepackage{makecell}
\usepackage{booktabs}
\usepackage{multirow}
\usepackage{color}
\usepackage{bm}

\usepackage{amsthm}
\usepackage{algorithm}
\usepackage{algorithmic}

\ifCLASSINFOpdf
\else
\fi
\begin{document}
%
%
%
%
\title{DEeR: Deviation Eliminating and Noise Regulating for Privacy-preserving Federated Low-rank Adaptation}
\author{Meilu Zhu, Axiu Mao, Jun Liu*, Yixuan Yuan*
\thanks{This work was supported by the Hong Kong Research Grants Council under Grant 11212321, 11217922, and ECS-21212720, the HKSAR Innovation and Technology Commission (ITC) under ITF Project MHP/109/19, ITS/229/22, and the Science, Technology and Innovation Committee of Shenzhen under Grant SGDX20210823104001011. (\textit{*Corresponding authors: Yixuan Yuan (yxyuan@ee.cuhk.edu.hk), Jun Liu (dr.jun.liu@hku.hk)})}
\thanks{
M. Zhu is with Department of Mechanical Engineering, City University of Hong Kong;
A. Mao is with School of Communication Engineering, Hangzhou Dianzi University, China;
J. Liu is with Department of Industrial and Manufacturing Systems Engineering, The University of Hong Kong and was with Department of Mechanical Engineering, City University of Hong Kong;
Y. Yuan is with Department of Electronic Engineering, Chinese University of Hong Kong and was with Department of Electrical Engineering, City University of Hong Kong.}}

\markboth{Journal of \LaTeX\ Class Files,~Vol.~14, No.~8, August~2015}%
{Shell \MakeLowercase{\textit{et al.}}: Bare Demo of IEEEtran.cls for IEEE Journals}
%



\maketitle

\begin{abstract}
Integrating low-rank adaptation (LoRA) with federated learning (FL) has received widespread attention recently, aiming to adapt pretrained foundation models (FMs) to downstream medical tasks via privacy-preserving decentralized training. However, owing to the direct combination of LoRA and FL, current methods generally undergo two problems, \textit{i.e.}, aggregation deviation, and differential privacy (DP) noise amplification effect. To address these problems, we propose a novel privacy-preserving federated finetuning framework called \underline{D}eviation \underline{E}liminating and Nois\underline{e} \underline{R}egulating (DEeR). Specifically, we firstly theoretically prove that the necessary condition to eliminate aggregation deviation is guaranteing the equivalence between LoRA parameters of clients. Based on the theoretical insight, a deviation eliminator is designed to utilize alternating minimization algorithm to iteratively optimize the zero-initialized and non-zero-initialized parameter matrices of LoRA, ensuring that aggregation deviation always be zeros during training.
Furthermore, we also conduct an in-depth analysis of the noise amplification effect and find that this problem is mainly caused by the ``linear relationship'' between DP noise and LoRA parameters. To suppress the noise amplification effect, we propose a noise regulator that exploits two regulator factors to decouple relationship between DP and LoRA, thereby achieving robust privacy protection and excellent finetuning performance.
Additionally, we perform comprehensive ablated experiments to verify the effectiveness of the deviation eliminator and noise regulator. DEeR shows better performance on public medical datasets in comparison with state-of-the-art approaches. The code is available at https://github.com/CUHK-AIM-Group/DEeR.
\end{abstract}

\section{Introduction}
\label{sec:introduction}
With the advent of the big data era and advances in computation~\cite{MENG2021100006}, large foundation models (FMs), such as CLIP~\cite{radford2021learning}, BiomedCLIP~\cite{zhang2023large}, SAM~\cite{kirillov2023segment}, have been developed, demonstrating unprecedented generalization performance across various medical tasks~\cite{qiu2023large, thirunavukarasu2023large}. However, these foundation models usually focus on general representation learning and still require further finetuning for downstream tasks~\cite{lin2023speciality, ma2024segment}. To avoid the computational burdens caused by finetuning entire foundation models,  various parameter-efficient finetuning (PEFT) methods~\cite{hu2021lora, jia2022visual,li2021prefix} have been proposed. One of the most widely used PEFT methods is low-rank adaptation (LoRA)~\cite{hu2021lora}, which adds a parallel branch of trainable adapters with parameters $\mathbf{A}$ and $\mathbf{B}$ to compute the model update $\nabla\mathbf{W}$. The ranks of $\mathbf{A}$ and $\mathbf{B}$ are much smaller than the pretrained model parameters $\mathbf{W}$. When applying LoRA for finetuning, only $\mathbf{A}$ and $\mathbf{B}$ are updated while the entire $\mathbf{W}$ is frozen, thereby significantly reducing GPU memory consumption~\cite{nguyen2024flora}.
\begin{figure}[!t]
\centering
\includegraphics[width = 0.5\textwidth]{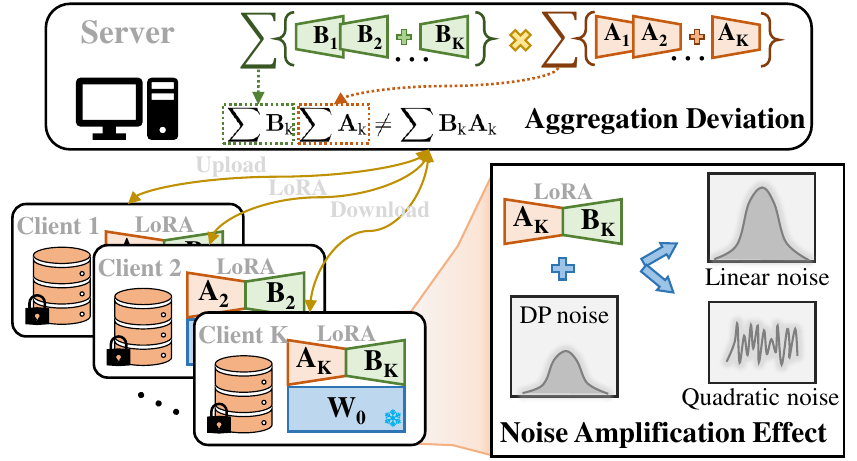}
\vspace{-4.0mm}
\caption{Directly combining FL and LoRA to finetune FMs brings two challenges, \textit{i.e.}, aggregation deviation and noise amplification effect.}
\label{fig:intro}
\vspace{-4.0mm}
\end{figure}

 Finetuning pretrained FMs with LoRA still requires sufficient training data for adaptation to specific downstream tasks~\cite{sun2024improving,babakniya2023slora}. Nevertheless, data within a single institution tend to be limited~\cite{FedOSS, FedDM}, particularly in medical scenarios. Directly gathering data from different institutions is typically unrealistic due to growing privacy concerns and legal restrictions~\cite{chen2021personalized,chen2022personalized,Yang_2023_ICCV}. An alternative approach is to adopt federated learning (FL)~\cite{mcmahan2017communication},  a decentralized learning paradigm, as training framework to collaboratively finetune FMs with LoRA. The FL paradigm~\cite{mcmahan2017communication, li2020federated} allows participating institutions (referred to as clients) to share their model gradients or parameters for the model aggregation at a trustworthy center (known as server), without leaking local raw data. Meanwhile, differential privacy (DP) techniques~\cite{dwork2006calibrating,kairouz2021advances, zhang2020batchcrypt} can be further employed to provide theoretical privacy guarantees against attacks and prevent local private information from being leaked during the communication process.

Recently, some methods~\cite{nguyen2024flora, wu2024fedlora, zhang2023fedpetuning, du2024communication, cho2023heterogeneous, babakniya2023slora, jiang2023low} have tried to integrate LoRA with FedAvg~\cite{mcmahan2017communication} in different applications. These methods finetune LoRA modules using local data of clients and then send the
updated modules to the server. The server averages all received LoRA modules to obtain a global LoRA, and distributes it to all clients as the initialization of the next round. Despite the promising performance, these approaches neglect two key issues, as shown in Fig.~{\ref{fig:intro}}. Firstly, the naive averaging of local LoRA modules leads to the \textbf{aggregation deviation} at the server side. In the FedAvg setting with LoRA, the local updates $\Delta \mathbf{W}$ of a client are decomposed into two low-rank matrices $\mathbf{A}$ and $\mathbf{B}$, $\Delta \mathbf{W} = \mathbf{B}\mathbf{A}$.  If $\mathbf{A}$ and $\mathbf{B}$ of all clients are aggregated independently, we will obtain a biased global update $\Delta \mathbf{W}_\textrm{g} = \sum \mathbf{B} \sum \mathbf{A} $. Theoretically, the true global update should be computed via $\Delta \mathbf{W}_\textrm{g}^{*} = \sum \Delta \mathbf{W} = \sum \mathbf{B} \mathbf{A} $. Obviously, there exists a mathematical deviation $\Delta \mathbf{W}_\textrm{g} \neq\Delta \mathbf{W}_\textrm{g}^{*} $, which would severely impede the convergence of a federation system.

The second problem, i.e., \textbf{noise amplification effect}, lies in the client side and arises from the intrinsic ``quadratic'' architecture of LoRA. Differential privacy (DP) is a commonly-used technique in FL to provide privacy guarantee by adding noise (e.g., Gaussian noise) to client gradients against training data leakage from the shared model~\cite{yang2023dynamic}. When the (Gaussian) noises $\bm{\xi}^{\mathbf{A}}$ and $\bm{\xi}^{\mathbf{B}}$ are injected into $\mathbf{A}$ and $\mathbf{B}$ of local LoRA modules, the ``quadratic'' architecture of LoRA would lead to the noise items $\mathbf{B}\bm{\xi}^{\mathbf{A}}$, $\bm{\xi}^{\mathbf{B}}\mathbf{A}$ and $\bm{\xi}^{\mathbf{B}}\bm{\xi}^{\mathbf{A}}$. In experiments, we observe that the noise intensities of the items $\mathbf{B}\bm{\xi}^{\mathbf{A}}$ and $\bm{\xi}^{\mathbf{B}}\mathbf{A}$ would be continuously amplified during training. The third term no longer follows a Gaussian distribution and also increases as the privacy budget of DP decreases. The noise amplification effect will hinder the model convergence when applying DP into a federated finetuning system with LoRA.

To overcome these problems, we propose a novel privacy-preserving federated finetuning (FedFT) framework called \underline{D}eviation \underline{E}liminating and Nois\underline{e} \underline{R}egulating (DEeR). The goal of DEeR is to adapt pretrained FMs to downstream medical tasks via LoRA in FL with client-level DP guarantees. Specifically, we firstly theoretically prove that the necessary condition to eliminate aggregation deviation is guaranteing the equivalence between LoRA parameters of clients.
With the theoretical insight, we design a deviation eliminator at the server side, which utilizes the alternating minimization algorithm to iteratively optimize the parameters $\mathbf{A}$ and $\mathbf{B}$ of LoRA, ensuring that aggregation deviation always be zeros during training. Moreover, we also conduct an in-depth analysis of the noise amplification effect and find that this problem is mainly caused by ``linear relationship'' between DP noise and LoRA parameters. To suppress the noise amplification effect, we propose a noise regulator that exploits two regulator factors to decouple the relationship between DP and LoRA, thereby achieving robust privacy protection and excellent finetuning performance.
The main contributions of this work are summarized as follows:
\begin{itemize}
    \item This work in-depth analyzes the challenges of a privacy-preserving FedFT system with LoRA. To the best of our knowledge, it represents the first effort to adapt different pretrained FMs to various downstream medical tasks via FedFT with LoRA.
    \item We propose a deviation eliminator that utilizes the alternating minimization algorithm to optimize the parameters of LoRA to avoid aggregation deviation.
    \item We present a noise regulator that can exploit two regulator factors to decouple relationship between DP and LoRA to suppress the noise amplification effect.
    \item Extensive experiments are conducted on public datasets. The results demonstrate the superior performance of the proposed DEeR over state-of-the-arts and the efficacy of different components.
\end{itemize}

\textbf{Roadmap.} The rest of the paper is organized as follows. In Section \ref{sec:related-work}, we review previous methods focusing on PEFT and federated finetuning with LoRA. Some preliminary knowledge is presented in Section \ref{sec:preliminaries}. In Section \ref{sec:method}, the proposed DEeR framework is introduced in detail. We describe implementation details, experimental settings and results in Section \ref{sec:experiments}. Finally, the paper is closed with the conclusion in Section \ref{sec:conclusion}.

\section{Related Work}
\label{sec:related-work}
We introduce existing methods about parameter efficient fine tuning and federated finetuning with LoRA in this section.
\subsection{Parameter Efficient Fine Tuning (PEFT)}

Parameter efficient fine tuning enables efficient adaptation of foundation models to various downstream tasks without the need to finetune all parameters of FMs. It only optimizes a small subset of parameters and thus results in significant reductions in computation and storage costs. Existing PEFT methods can be broadly divided into three main categories.

The first category is dedicated to designing task-related \underline{Adapters}~\cite{sung2022vl, xu2023side}. For example, VL-Adapter~\cite{sung2022vl} inserts trainable adapter modules into a fixed CLIP model and finetunes only the adapters for vision-language tasks. SAN~\cite{xu2023side} presents a decoupled structure to reduce computational costs for semantic segmentation, \textit{i.e.}, introducing an adapter network as the side branch of FMs.
\underline{Prompt tuning} falls into the second category. The prompt tuning~\cite{zhou2022learning} originally treats the prompts in NLP as task-specific continuous vectors and only optimizes
them during finetuning, while visual prompt tuning~\cite{jia2022visual} uses a set of continuous embeddings as visual prompts to pad the patch embeddings. However, both Adapter and Prompt tuning-based approaches introduce extra parameters and result in inference latency~\cite{han2024parameter}.

To solve this problem, the third type of works focus on reparameterization techniques, the most famous of which is the \underline{LoRA series}~\cite{han2024parameter}. The vanilla LoRA~\cite{hu2021lora}
optimizes rank decomposition matrices and re-parameterize the pretrained weight matrices. Numerous studies~\cite{lin2024lora,zhang2023adaptive, zhong2024convolution,wu2024mole, liu2023moelora} have further developed and applied LoRA to various scenarios. For instance, considering that prespecifing a rank for all layers neglects the importance of different layers, AdaLoRA~\cite{zhang2023adaptive} dynamically allocates the rank for different layers by importance scoring. Additionally, LoRA Dropout~\cite{zhang2023adaptive} observes that finetuning LoRA-series models also face the risk of overfitting and thus introduces dropout technique to randomly drop rows and columns from tunable low-rank parameter matrices. Recent initiatives~\cite{wu2024mole,liu2023moelora} mainly focus on the composition of separate trained LoRAs to amplify performance across various tasks. For example, MOLE~\cite{wu2024mole} treats each layer of trained LoRAs as a distinct expert and learns a gating function to get composition weights to fuse these experts. Differing to existing approaches \cite{lin2024lora,zhang2023adaptive, zhong2024convolution,wu2024mole, liu2023moelora}, this paper represents the first effort to develop LoRA into medical domain in a decentralized learning setting, aiming to achieve privacy-preserving federated PEFT.

\subsection{Federated Finetuning with LoRA}
The above PEFT approaches generally assume that training data comes from a data warehouse. In practice, however, data is often owned by multiple parties and is often prohibited from being shared with others, especially in medical domain. Recently, interest in the intersection of PEFT and FL has notably increased, forming a new research topic called FedFT. FedPETuning~\cite{zhang2023fedpetuning} provides a holistic empirical study of representative PEFT methods in FL. The experimental results show that the LoRA-based FedFT technique achieves a very promising performance and inference speed. Next, we review the existing  LoRA-based FedFT methods.

Among current LoRA-based FedFT methods, a common solution is directly combining LoRA with FL to finetune FMs for various scenes, such as speech-to-text tasks~\cite{du2024communication} and personalized FL~\cite{wu2024fedlora}.
Nevertheless, this native way presents slow a convergence speed and leads to costly communication expenses. SLoRA~\cite{babakniya2023slora} and FeDeRA~\cite{yan2024federa}
attribute the problem to random initialization of low-rank matrices. To this end, they use the singular value decomposition (SVD) to obtain better initialization from the pretrained full matrix.
Besides, there is another problem that different layers of all client models should share varying ranks due to heterogeneous resources and data distributions. SA-FedLora~\cite{yang2024safedlora} defines a scheduler function to adaptively adjust rank with communication round. In HETLORA~\cite{cho2023heterogeneous}, all clients first start from a global rank and then self-prune their respective ranks based on the magnitude of the model parameters. FlexLoRA~\cite{yan2024federa} encourages clients to use different ranks during local training and upload full-size LoRA to a server. The server uses SVD to decouple the aggregated full-size matrix and distributes different sizes of LoRA to the clients. thereby achieving heterogeneous LoRA.

Apart from the above problems, FFA-LoRA~\cite{sun2024improving} finds that the ``quadratic'' structure of LoRA incurs aggregation bias and introduces quadratic DP noise. To break the ``quadratic'' structure, FFA-LoRA fixes the randomly initialized non-zero matrices and only finetunes the zero-initialized matrices.
However, freezing non-zero matrices will hinder the model from converging to a good local minimum, since random initialization is nearly impossible to produce optimal parameters for downstream tasks~\cite{glorot2010understanding, he2015delving}. This strategy makes FFA-LoRA very sensitive to different initialization. A bad initialization can degrade the model performance. Experiments in the previous method~\cite{kuo2024federated} and our paper also show the limited performance of FFA-LoRA. In addition, FFA-LoRA neglects the effect of linear noises. In this paper, we provide a more comprehensive study about these issues and an in-depth analyze the conditions to solve these problems.

\section{Preliminaries}
\label{sec:preliminaries}
In this section, we present some background knowledge about LoRA, federated finetuning (FedFT) with LoRA, and differential privacy.

\textbf{LoRA}. As one of the most promising PEFT methods in the central setting, the key idea of LoRA~\cite{hu2021lora} is decomposing the update $\Delta\mathbf{W} \in \mathbb{R}^{m \times n}$ of target module into low-rank matrices:
\begin{equation}\label{eq1}
\begin{aligned}
\mathbf{W}_0 + \Delta \mathbf{W} = \mathbf{W}_0 + \mathbf{B}\mathbf{A},
\end{aligned}
\end{equation}
where $\mathbf{W}_0 \in \mathbb{R}^{m \times n}$ denotes the pretrained weight matrix. $\mathbf{B} \in \mathbb{R}^{m \times r}$ and $\mathbf{A} \in \mathbb{R}^{r \times n}$ are the low-rank decomposition of $\Delta\mathbf{W}$, such that $\Delta \mathbf{W} = \mathbf{B}\mathbf{A}$. Typically, $r$ is the rank of $\Delta \mathbf{W}$, $\mathbf{B}$, $\mathbf{A}$, and significantly smaller than $m$ and $n$.
During the finetuning phase, the model optimizes matrices $\mathbf{B}$ and $\mathbf{A}$ instead of directly updating $\mathbf{W}_0$, thus achieving the substantial reduction in
GPU memory and storage usage. Additionally, to ensure the stable convergence, $\mathbf{B}$ and $\mathbf{A}$ use zero and random Gaussian initialization\emph{} respectively, so that $\Delta \mathbf{W} = \mathbf{B}\mathbf{A}$ is zero at the beginning of training.

\textbf{FedFT with LoRA}. Current LoRA-based FedFT methods \cite{nguyen2024flora, wu2024fedlora, zhang2023fedpetuning, du2024communication, cho2023heterogeneous, jiang2023low} follow a standard FL setting, \textit{i.e.}, FedAvg~\cite{mcmahan2017communication}. These methods collaboratively unite local LoRA modules of $K$ clients to learn a global LoRA ($\mathbf{B}_\textrm{g}$, $\mathbf{A}_\textrm{g}$) as the global change $\Delta \mathbf{W}_\textrm{g}$, enabling the pretrained knowledge $\mathbf{W}_\textrm{0}$ to adapt downstream tasks via multiple rounds of communication:
\begin{equation}\label{eq2}
\begin{aligned}
\mathbf{W}_\textrm{g} = \mathbf{W}_\textrm{0}+\Delta \mathbf{W}_\textrm{g} = \mathbf{W}_\textrm{0}+\mathbf{B}_\textrm{g}\mathbf{A}_\textrm{g},
\end{aligned}
\end{equation}
where $\mathbf{B}_\textrm{g}$ and $\mathbf{A}_\textrm{g}$ are obtained via the aggregation of local LoRA modules as follows,
\begin{equation}\label{eq3}
\mathbf{B}_\textrm{g} = 1/K\sum\nolimits_{k\in [K]}\mathbf{B}_k, \ \ \ \mathbf{A}_\textrm{g} = 1/K\sum\nolimits_{k\in [K]}\mathbf{A}_k.
\end{equation}
The updated $\mathbf{B}_\textrm{g}$ and $\mathbf{A}_\textrm{g}$ are distributed back to clients as the initialization of local LoRA modules in the next round.

\textbf{Differential Privacy}
Differential privacy (DP) is a popular manner to provide theoretical guarantees against training data leakage from the model in federated learning~\cite{dwork2006calibrating}. This work focuses on the client-level DP, aiming to ensure information security for any clients.
\theoremstyle{definition}
\newtheorem{definition}{Definition}
\begin{definition}
(Client-level DP) \textit{A randomized algorithm $\mathcal{M}$ is $(\varepsilon, \delta)$-DP if for any two adjacent datasets $\mathcal{D}$, $\mathcal{D}'$ constructed by adding or removing all
 records of any client, and every possible subset of outputs $\mathcal{S}$ in the range of $\mathcal{M}$ satisfy the following inequality:}
\begin{equation}\label{eq31}
\mathrm{Pr}[\mathcal{M}(\mathcal{D})\in \mathcal{S}] \leq e^\varepsilon \mathrm{Pr}[\mathcal{M}(\mathcal{D}')\in \mathcal{S}]+\delta.
\end{equation}
where the parameter $\varepsilon$ is called the privacy budget and a smaller $\varepsilon$  means a stronger privacy protection guarantee. The parameter $\delta$ defines
the probability of failing to guarantee the differential privacy bound for any two adjacent datasets. At each round, each client first clips the local gradient with a norm constraint $C$. After clipping, we add Gaussian noise to the gradient before uploading it to the server~\cite{shi2023make}, as follows:
\begin{equation}\label{eq32}
\Delta \mathbf{W} = \Delta \mathbf{W}*\min(1, \frac{C}{\|\Delta \mathbf{W}\|_2}) + \mathcal{N}(0, \sigma^2C^2\cdot \mathbf{I}_d/K)
\end{equation}
where $\sigma$ is noise variance.
\end{definition}

\section{methodology}
\label{sec:method}
\begin{figure}[!t]
\centering
\includegraphics[width = 0.5\textwidth]{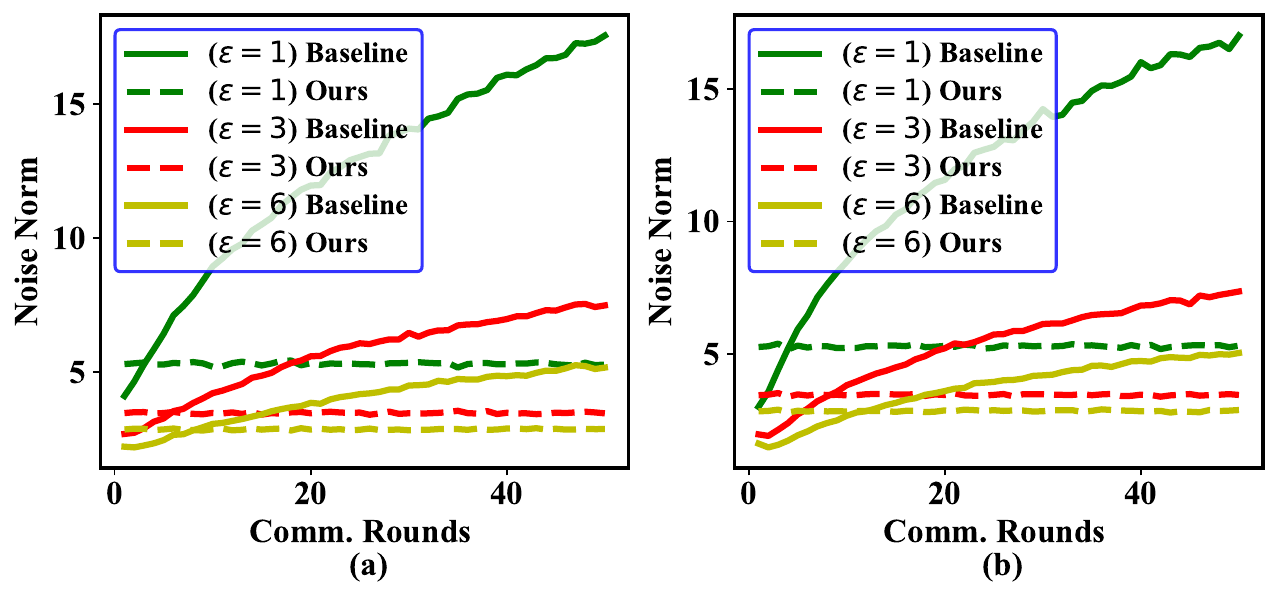}
\vspace{-5.0mm}
\caption{The norm changes of linear noise items with communication round: (a) $\|\bm{\xi}_k^{\mathbf{B}}\mathbf{A}_k\|_F$, (b)$\|\mathbf{B}_k\bm{\xi}_k^{\mathbf{A}}\|_F$. (Best viewed in color.)}
\label{fig:noise}
\vspace{-3.0mm}
\end{figure}
We first exhaustively analyze the challenges faced by FedFT with LoRA. Then, the overall framework and its submodules are introduced.
\subsection{Problem Formulation}
The native way to integrate LoRA and FedAvg in existing FedFT methods~\cite{nguyen2024flora, wu2024fedlora, zhang2023fedpetuning, du2024communication, cho2023heterogeneous, jiang2023low} neglects the impact of ``quadratic'' architecture of LoRA. It would incur two intractable issues and hinder convergence of a federation system~\cite{sun2024improving}.

\textbf{Aggregation Deviation}. The first issue is aggregation deviation at the server side. Theoretically, the expected global model update $\Delta \mathbf{W}_\textrm{g}$ in Eq. (\ref{eq2}) should be calculated by averaging the uploaded updates $\{\Delta \mathbf{W}_k\}^K_{k=1}$ of clients~\cite{mcmahan2017communication}:
\begin{equation}\label{eq4}
\Delta \mathbf{W}_\textrm{g} = \frac{1}{K}\sum\nolimits_{k\in [K]} \Delta \mathbf{W}_k = \frac{1}{K}\sum\nolimits_{k\in [K]} \mathbf{B}_k\mathbf{A}_k.
\end{equation}
However, according to Eq. (\ref{eq3}), existing approaches~\cite{nguyen2024flora, wu2024fedlora, zhang2023fedpetuning, du2024communication, cho2023heterogeneous, jiang2023low} aggregate $\mathbf{B}$ and $\mathbf{A}$ parts separately. Obviously, there exists an aggregation deviation as follows:
\begin{equation}\label{eq5}
\small
\Delta \mathbf{W}_\textrm{g} = \frac{1}{K}\sum\nolimits_{k\in [K]} \mathbf{B}_k \frac{1}{K}\sum\nolimits_{k\in [K]}\mathbf{A}_k \neq \frac{1}{K}\sum\nolimits_{k\in [K]} \mathbf{B}_k \mathbf{A}_k.
\end{equation}
Essentially, the deviation comes from the contradiction between model averaging in FL and the ``quadratic'' architecture of LoRA.
In next section, we theoretically analyze that the deviation becomes larger when the data of clients are more heterogeneous.

\textbf{Noise Amplification Effect}. Another problem is that the ``quadratic'' architecture of LoRA would amplify DP noise. For the convenience of discussion, we omit the gradient clipping step and focus on one round of training. Generally, after local training of client $k$, the noise $\bm{\xi}_k \in \mathbb{R}^{m\times n}$ is sampled from Gaussian distribution and added to the local update $\Delta \mathbf{W}_k\in \mathbb{R}^{m\times n}$ to ensure client-level DP as follows:
\begin{equation}\label{eq6}
\widetilde{\mathbf{W}}_k = \mathbf{W}_k + \bm{\xi}_k =  \mathbf{W}_0 + (\Delta \mathbf{W}_k + \bm{\xi}_k),
\end{equation}
where $\widetilde{\mathbf{W}}_k$ is the local model after adding noise.
However, in FedTF with LoRA, we exploit LoRA to replace $\Delta \mathbf{W}_k$ and upload parameters $\mathbf{B} $ and $\mathbf{A}$ to the server. Therefore, we need to add Gaussian noise $\bm{\xi}_k^{\mathbf{B}}\in \mathbb{R}^{m\times r}$ and $\bm{\xi}_k^{\mathbf{A}}\in \mathbb{R}^{r\times n}$ to $\mathbf{B}_k $ and $\mathbf{A}_k$ instead of $\Delta \mathbf{W}_k$:
\begin{equation}\label{eq7}
\begin{aligned}
\widetilde{\mathbf{W}}_k &= \mathbf{W}_0 + (\mathbf{B}_k + \bm{\xi}_k^{\mathbf{B}})(\mathbf{A}_k + \bm{\xi}_k^{\mathbf{A}}) \\
&=\mathbf{W}_0 + \mathbf{B}_k\mathbf{A}_k + \mathbf{B}_k\bm{\xi}_k^{\mathbf{A}}   + \bm{\xi}_k^{\mathbf{B}}\mathbf{A}_k  + \bm{\xi}_k^{\mathbf{B}}\bm{\xi}_k^{\mathbf{A}},
\end{aligned}
\end{equation}
where we call the third and fourth terms as \underline{linear noises}, the final term as \underline{quadratic noise}. FFA-LoRA~\cite{sun2024improving} has shown that the quadratic noise becomes larger with the smaller privacy budget $\varepsilon$ and hinders the convergence of the federated system.

Yet, FFA-LoRA neglects the impact of linear noises. To reveal the characteristics of linear noises $\bm{\xi}_k^{\mathbf{B}}\mathbf{A}_k$ and $\mathbf{B}_k\bm{\xi}_k^{\mathbf{A}}$, we implement a baseline version of FedFT with LoRA~\cite{nguyen2024flora, wu2024fedlora, zhang2023fedpetuning, du2024communication, cho2023heterogeneous, jiang2023low} on an endoscopic dataset, \textit{i.e}, Kvasir-v2~\cite{Pogorelov2017kvasir}, where the number of clients is 12 and the pretrained model is BiomedCLIP~\cite{zhang2023large}. We randomly select one layer of LoRA in a client model and plot the Frobenius norm changes of its linear noises with the communication round, under different privacy budgets $\varepsilon \in \{1, 3, 6\}$, as shown in Fig.~\ref{fig:noise}~(a)~and~(b). We can observe that: (1) The noise norms $\|\mathbf{B}_k\bm{\xi}_k^{\mathbf{A}}\|_F$ and $\|\bm{\xi}_k^{\mathbf{B}}\mathbf{A}_k\|_F$ continuously increase with the communication round for any privacy budgets; (2) The increase rates of $\|\mathbf{B}_k\bm{\xi}_k^{\mathbf{A}}\|_F$ and $\|\bm{\xi}_k^{\mathbf{B}}\mathbf{A}_k\|_F$ are greater when the privacy budget $\varepsilon$ becomes smaller. These observations
confirm that the ``quadratic'' architecture of LoRA can enlarge the original DP noise for a given privacy budget. This leads to the privacy guarantee shrinking since we need to increase the privacy budget to ensure model convergence.

\begin{algorithm}[!t]
    \label{algorithm1}
    \footnotesize
    \caption{The proposed DEeR algorithm for federated low-rank adaptation.}
    \begin{algorithmic}[1]
        \STATE \textbf{Server executes:}
        \FOR{each communication round $t$}
        \FOR{each client $k = 1,2,...,K$}
        \STATE Downloading $\mathbf{A}_\textrm{g}^{(t)}$ to update $\mathbf{A}_k^{(t-1)}$ to $\mathbf{A}_k^{(t)}$ and freezing it.
        \STATE $\mathbf{B}_k^{(t+1)}\leftarrow$ ClientUpdate\_B($\mathbf{W}_0, \mathbf{A}_\textrm{g}^{(t)}, \mathbf{B}_k^{(t)}$).
        \ENDFOR
        \STATE Aggregating $\mathbf{B}$:  $\mathbf{B}_\textrm{g}^{(t+1)}\leftarrow \sum_{k=1}^K \mathbf{B}_k^{(t+1)} $.
        \FOR{each client $k = 1,2,...,K$}
        \STATE Downloading $\mathbf{B}_\textrm{g}^{(t+1)}$ to update $\mathbf{B}_k^{(t)}$ to $\mathbf{B}_k^{(t+1)}$ and freezing it.
        \STATE $\mathbf{A}_k^{(t+1)}\leftarrow$ ClientUpdate\_A($\mathbf{W}_0, \mathbf{A}_k^{(t)}, \mathbf{B}_\textrm{g}^{(t+1)}$).
        \ENDFOR
        \STATE Aggregating $\mathbf{A}$:  $\mathbf{A}_\textrm{g}^{(t+1)}\leftarrow \sum_{k=1}^K \mathbf{A}_k^{(t+1)} $.
        \ENDFOR
        \STATE \textbf{Client executes:}
        \STATE ClientUpdate\_B($\mathbf{W}_0, \mathbf{A}_\textrm{g}^{(t)}, \mathbf{B}_k^{(t)}$):
        \STATE ~~~~\textbf{for} {each epoch $e = 1,2,...,E$} \textbf{do}
        \STATE ~~~~~~~~$\mathbf{B}_k^{(t+1)}\leftarrow \arg\min_{\mathbf{B}_k} f_k(\mathbf{W}_0, \mathbf{A}_\textrm{g}^{(t)},\mathbf{B}_k^{(t)})$.
        \STATE ~~~~\textbf{end for}
        \STATE ~~~~ Based on Theorem 2, adding the modulated Gaussian noise into $\mathbf{B}_k^{(t+1)}$ before uploading it.
        \STATE ClientUpdate\_A($\mathbf{W}_0, \mathbf{A}_k^{(t)}, \mathbf{B}_\textrm{g}^{(t+1)}$):
        \STATE ~~~~\textbf{for} {each epoch $e = 1,2,...,E$} \textbf{do}
        \STATE ~~~~~~~~$\mathbf{A}_k^{(t+1)}\leftarrow \arg\min_{\mathbf{A}_k} f_k(\mathbf{W}_0, \mathbf{A}_k^{(t)},\mathbf{B}^{(t+1)}_\textrm{g})$.
        \STATE ~~~~\textbf{end for}
        \STATE ~~~~Based on Theorem 2, adding the modulated Gaussian noise into $\mathbf{A}_k^{(t+1)}$ before uploading it.
    \end{algorithmic}
    \label{algorithm1}
\end{algorithm}
\begin{figure*}[!t]
\centering
\includegraphics[width = 1\textwidth]{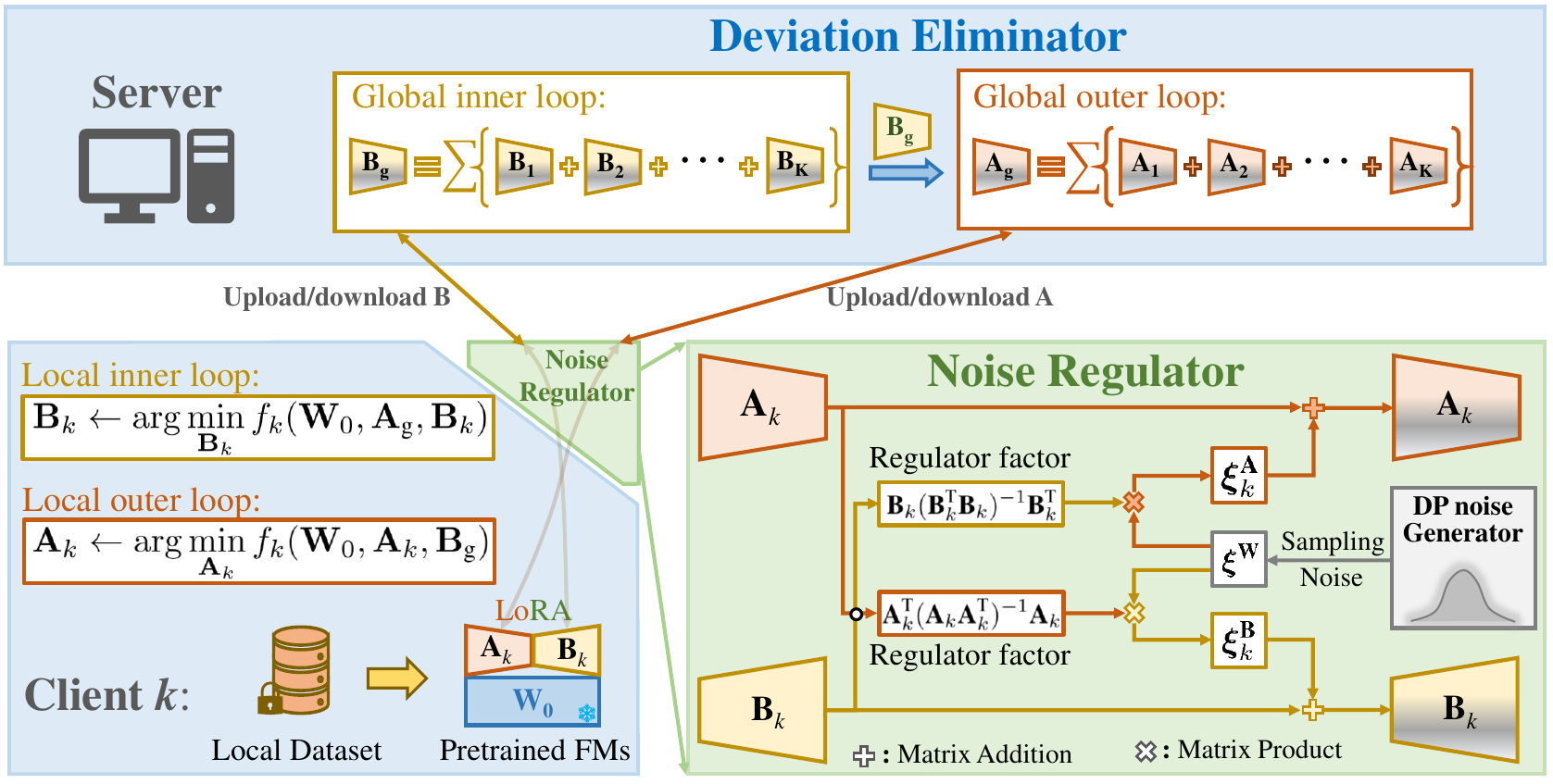}
\caption{The overview of the proposed DEeR framework for federated finetuning with LoRA (Best viewed in color). DEeR is equipped with a deviation eliminator at the server side and a noise regulator at the client side. The deviation eliminator exploits alternating minimization algorithm to optimize the parameters $\mathbf{A}$ and $\mathbf{B}$ of LoRA for mitigating aggregation deviation. The noise regulator introduces two regulator factors to  suppress noise amplification effect.}
\label{fig:framework}
\vspace{-2.0mm}
\end{figure*}
\subsection{Overview}
We present \underline{D}eviation \underline{E}liminator and Nois\underline{e} \underline{R}egulator (DEeR), a privacy-preserving federated finetuning framework to adapt pretrained foundation models to downstream medical tasks via LoRA. Similar to previous works~\cite{nguyen2024flora, wu2024fedlora, zhang2023fedpetuning, du2024communication, cho2023heterogeneous, jiang2023low},  DEeR follows the standard FL setting, \textit{i.e.}, FedAvg~\cite{mcmahan2017communication}, and collaborates $K$ clients to finetune a pretrained foundation model with the frozen parameters $\mathbf{W}_0$ and trainable LoRA parameters $\mathbf{A}$ and $\mathbf{B}$, as shown in Fig.~\ref{fig:framework}. Each client holds a local medical dataset $\mathcal{D}^k = \{(\mathbf{x}_i^k,  \mathbf{y}_i^k)\}$, where $\mathbf{x}_i^k$ denotes a training sample with the label $\mathbf{y}_i^k$. At the beginning of training, the server sends the parameters $\mathbf{W}_0$, $\mathbf{A}$ and $\mathbf{B}$
to clients. During the overall training process, $\mathbf{W}_0$ is fixed and only the parameters $\mathbf{A}$ and $\mathbf{B}$ are updated and uploaded to the server for aggregation. $\mathbf{A}$ and $\mathbf{B}$ have low communication costs since they are low-rank matrices. DEeR equips every client with a noise regulator to suppress the noise amplification effect caused by the ``quadratic'' architecture of LoRA during local training. Meanwhile, a deviation eliminator on the server side is used to schedule the optimization process of $\mathbf{A}$ and $\mathbf{B}$ to avoid the aggregation deviation. In addition, we provide the pseudocode to show the workflow of DEeR in Algorithm~\ref{algorithm1}.

\subsection{Deviation Eliminator} 
Although LoRA is favored by FedFT since its low-rank property and ``quadratic'' architecture result in the low communication costs and relatively low computational burdens, it also introduces the new challenge \textit{i.e.}, aggregation deviation, as shown in Eq~(\ref{eq5}).
To tackle this dilemma, we present a deviation eliminator that exploits alternating minimization algorithm to decouple the ``quadratic'' architecture of LoRA and achieve the robust optimization of the parameters $\mathbf{A}$ and $\mathbf{B}$, as demonstrated in Fig.~\ref{fig:framework}.

To quantify the overall deviation $\mathcal{O}$ owing to LoRA aggregation on the server side, we define the deviation term based on Eq.~(\ref{eq5}) as below:
\begin{equation}\label{eq11}
\mathcal{O} = \left|\frac{1}{K^2}\sum\nolimits_{k\in [K]} \mathbf{B}_k \sum\nolimits_{k\in [K]}\mathbf{A}_k - \frac{1}{K}\sum\nolimits_{k\in [K]} \mathbf{B}_k \mathbf{A}_k\right|.
\end{equation}
We theoretically show that the overall deviation $\mathcal{O}$ can be eliminated when $\mathbf{A}$ or $\mathbf{B}$ of all clients are equivalent.
\theoremstyle{theorem}
\newtheorem{theorem}{Theorem}
\begin{theorem} \label{theorem1}
Given a collection of $K$ clients, let $\mathbf{B}_k$, $\mathbf{A}_k$ and $\mathbf{B}_{k'}$, $\mathbf{A}_{k'}$ be the LoRA parameters of any two clients $k$ and $k'$, respectively. The overall aggregation deviation $\mathcal{O}$ will be zero when $\mathbf{B}_k$ and $\mathbf{B}_{k'}$ or $\mathbf{A}_k$ and $\mathbf{A}_{k'}$ are equivalent in a FedTF system with LoRA.
\end{theorem}
\noindent\textit{Proof of Theorem 1}.
\begin{equation}\label{eq12}
\small
\begin{aligned}
\mathcal{O} &= \left|\frac{1}{K^2}\sum\nolimits_{k\in [K]} \mathbf{B}_k\sum\nolimits_{k\in [K]}\mathbf{A}_k - \frac{1}{K}\sum\nolimits_{k\in [K]} \mathbf{B}_k \mathbf{A}_k\right| \\
&=\left|\frac{1}{K^2}(\sum\nolimits_{k\in [K]} \mathbf{B}_k\sum\nolimits_{k\in [K]}\mathbf{A}_k- K \sum\nolimits_{k\in [K]} \mathbf{B}_k \mathbf{A}_k)\right|\\
&=\left|\frac{1}{K^2}\left[ \sum\nolimits_{k'\in [K]} \left(\mathbf{B}_{k'}\sum\nolimits_{k\in [K]}\mathbf{A}_k-\sum\nolimits_{k\in [K]} \mathbf{B}_k \mathbf{A}_k\right) \right]\right| \\
&=\left|\frac{1}{K^2}\left[ \sum\nolimits_{k'\in [K]} \left(\sum\nolimits_{k\in [K]}\mathbf{B}_{k'}\mathbf{A}_k-\sum\nolimits_{k\in [K]} \mathbf{B}_k \mathbf{A}_k\right) \right]\right| \\
&=\left|\frac{1}{K^2}\left[ \sum\nolimits_{k'\in [K]}\ \ \sum\nolimits_{k\in [K]} (\mathbf{B}_{k'}-\mathbf{B}_{k})\mathbf{A}_k \right]\right| \\
&\ \ \ //\textmd{Using the similar derivation process.}// \\
&=\left|\frac{1}{K^2}\left[ \sum\nolimits_{k\in [K]}\ \ \sum\nolimits_{k'\in [K]} \mathbf{B}_{k}(\mathbf{A}_{k'}-\mathbf{A}_k) \right]\right|. \ \ \ \ \ \ \ \ \ \ \ \  \ \square  \\
\end{aligned}
\nonumber
\end{equation}
We can find that the deviation $\mathcal{O}$ will be zero when $\mathbf{B}$ or $\mathbf{A}$  of all clients are equivalent. Additionally, data heterogeneity has a significant impact on $\mathcal{O}$, since it increases the divergence between ($\mathbf{A}$, $\mathbf{B}$) of different clients.

With the insight of \textbf{Theorem \ref{theorem1}}, we introduce a constraint condition into the optimization objective of a FedFT system to ensure the equivalence between $\mathbf{A}$ or $\mathbf{B}$ of any two clients:
\begin{equation}\label{eq13}
\begin{aligned}
\min_{\mathbf{A}_\textrm{g},\mathbf{B}_\textrm{g}} & \frac{1}{K}\sum\nolimits_{k\in [K]} f_k(\mathbf{W}_0, \mathbf{A}_\textrm{g},\mathbf{B}_\textrm{g}),  \\
 \textrm{s.t.} &\sum\nolimits_{k\in [K]}\sum\nolimits_{k'\in [K]} (\mathbf{B}_{k}-\mathbf{B}_{k'})(\mathbf{A}_{k}-\mathbf{A}_{k'})=0,
\end{aligned}
\end{equation}
where $f_k$ is the loss function of the client $k$. The global $\mathbf{A}_\textrm{g}$ and $\mathbf{B}_\textrm{g}$ are obtained by Eq.~(\ref{eq3}).
The objective in Eq.~(\ref{eq13}) is equivalent to minimizing the following loss function for any client $k$ as follows:
\begin{equation}\label{eq14}
\small
\min_{\mathbf{A}_k,\mathbf{B}_k} f_k(\mathbf{W}_0, \mathbf{A}_k,\mathbf{B}_k),  \textrm{s.t.}, \sum_{k'\in [K]} (\mathbf{B}_{k}-\mathbf{B}_{k'})(\mathbf{A}_{k}-\mathbf{A}_{k'})=0.
\end{equation}
Eq.~(\ref{eq14}) cannot be directly optimized because ($\mathbf{A}$, $\mathbf{B}$) of each client are unavailable to each other during local training. To enable the optimization objective to be separable across clients, we employ the generalized Alternating Minimization (gAM) optimization algorithm~\cite{jain2017non}.
Specifically, at the $t$-th round, we firstly update ($\mathbf{A}_k^{(t-1)}$, $\mathbf{B}_k^{(t-1)}$) to ($\mathbf{A}_k^{(t)}$, $\mathbf{B}_k^{(t)}$) using the global ($\mathbf{A}_\textrm{g}^{(t)}$, $\mathbf{B}_\textrm{g}^{(t)}$) downloaded from the server, and then freeze $\mathbf{A}_k^{(t)}$ and only optimize $\mathbf{B}_k^{(t)}$ as follows:
\begin{equation}\label{eq15}
\mathbf{B}_k^{(t+1)}\leftarrow \arg\min_{\mathbf{B}_k} f_k(\mathbf{W}_0, \mathbf{A}_\textrm{g}^{(t)},\mathbf{B}_k^{(t)}).
\end{equation}
Since $\mathbf{A}^{(t)}$ of all clients equal to $\mathbf{A}_\textrm{g}^{(t)}$, the constraint condition are satisfied in Eq.~(\ref{eq14}). Next, the obtained $\{\mathbf{B}_k^{(t+1)}\}^K_{k=1}$ are delivered to the server and aggregated to get $\mathbf{B}^{(t+1)}_\textrm{g}$, which is distributed back to all clients to update $\mathbf{B}_k^{(t)}$ to $\mathbf{B}_k^{(t+1)}$. We fix $\mathbf{B}_k^{(t+1)}$ and only optimize $\mathbf{A}_\textrm{k}^{(t)}$ as follows:
\begin{equation}\label{eq16}
\mathbf{A}_k^{(t+1)}\leftarrow \arg\min_{\mathbf{A}_k} f_k(\mathbf{W}_0, \mathbf{A}_k^{(t)},\mathbf{B}^{(t+1)}_\textrm{g}).
\end{equation}
Similarly, $\mathbf{B}^{(t+1)}$ of all clients are  equivalent, so the constraint condition is also satisfied in Eq.~(\ref{eq14}).

The proposed deviation eliminator can exploit the gAM optimization algorithm to ensure the equivalence between $\mathbf{A}$ or $\mathbf{B}$ of any two clients for each round of aggregation. Therefore, the overall aggregation deviation $\mathcal{O}$ is always zero, thereby guaranteing the stable convergence of the model. Notably, FFA-LoRA~\cite{sun2024improving} fixes randomly-initialized $\mathbf{A}$ and only optimizes $\mathbf{B}$ during the overall training process, which can be regarded as a special case of our method. However, randomly-initialized $\textbf{A}$ is not certainly optimal and limits model convergence. In contrast, our approach optimizes $\mathbf{A}$ and $\mathbf{B}$ through gAM algorithm and is therefore more likely to converge to a better local optimum.

\subsection{Noise Regulator} 
Differential privacy (DP) techniques provide more stringent privacy protection to a FedFT system against potential privacy leaks. However, when directly introducing DP into a LoRA-based FedFT system, the ``quadratic'' architecture of LoRA would amplify DP noise, showing a significant negative impact on the model convergence and final finetuning performance. To suppress the noise amplification effect, we propose a novel noise regulator that uses two regulator factors to decouple the relationship between DP and LoRA, thereby ensuring robust privacy protection while maintaining superior finetuning performance, as illustrated in Fig.~\ref{fig:framework}.

As we discussed earlier about the noise amplification effect, LoRA transfers the original DP noise into two types, \textit{i.e.},  linear noises and quadratic noise, as demonstrated in Eq.~(\ref{eq7}). Because the ``quadratic'' structure of LoRA has been decoupled by the proposed deviation eliminator, the quadratic noise is eliminated~\cite{sun2024improving}. Concretely, after optimizing $\mathbf{A}$ and $\mathbf{B}$ via Eq.~(\ref{eq15}) and Eq.~(\ref{eq16}) respectively, we inject DP noise into them before sending them to the server as follows:
\begin{equation}\label{eq171}
\begin{aligned}
\mathbf{W}_0 + (\mathbf{B}_k + \bm{\xi}_k^{\mathbf{B}})\mathbf{A}_k =  \mathbf{W}_0 + \mathbf{B}_k\mathbf{A}_k + \bm{\xi}_k^{\mathbf{B}}\mathbf{A}_k,
\end{aligned}
\end{equation}
\begin{equation}\label{eq172}
\begin{aligned}
\mathbf{W}_0 + \mathbf{B}_k(\mathbf{A}_k + \bm{\xi}_k^{\mathbf{A}})=\mathbf{W}_0 + \mathbf{B}_k\mathbf{A}_k + \mathbf{B}_k\bm{\xi}_k^{\mathbf{A}},
\end{aligned}
\end{equation}
where we omit the superscript $t$ for convenience. In Eq.~(\ref{eq171}), $\mathbf{A}_k =\mathbf{A}_\textrm{g}$, and $\mathbf{B}_k =\mathbf{B}_\textrm{g}$ in Eq.~(\ref{eq172}). Although the quadratic noise disappears, the linear noises $\bm{\xi}_k^{\mathbf{B}}\mathbf{A}_k$ and $\mathbf{B}_k\bm{\xi}_k^{\mathbf{A}}$ are amplified with the communication round and still affect the learning of $\mathbf{A}_k$ and $\mathbf{B}_k$.

Next, we conduct an in-depth analysis of $\bm{\xi}_k^{\mathbf{B}}\mathbf{A}_k$ and  $\mathbf{B}_k\bm{\xi}_k^{\mathbf{A}}$. With a given privacy budget $\varepsilon$, we can obtain the corresponding Gaussian noise distribution by privacy composition rules~\cite{abadi2016deep}. Theoretically, if we sample DP noises $\bm{\xi}_k^{\mathbf{B}}\in \mathbb{R}^{m\times r}$ and $\bm{\xi}_k^{\mathbf{A}}\in \mathbb{R}^{r\times n}$ from the distribution, their norm values do not change significantly during the training process. Hence, the amplification of $\bm{\xi}_k^{\mathbf{B}}\mathbf{A}_k$ and $\mathbf{B}_k\bm{\xi}_k^{\mathbf{A}}$ is only related to $\mathbf{A}_k$ and $\mathbf{B}_k$, respectively. Next, we theoretically prove that the amplification effect can be removed by introducing two regulator factors.
\begin{theorem} \label{theorem2}
Assuming that $\mathbf{B}_k\in \mathbb{R}^{m\times r} $ and $\mathbf{A}_k\in \mathbb{R}^{r\times n} $ are LoRA parameters of the client $k$. Let $\bm{\xi}^{\mathbf{W}}\in \mathbb{R}^{m\times n} $ be DP noise sampled from a Gaussian distribution. $\mathbf{A}_k^{\textrm{T}}(\mathbf{A}_k\mathbf{A}_k^{\textrm{T}})^{-1}$ and $({\mathbf{B}_k^\textrm{T}\mathbf{B}_k})^{-1}\mathbf{B}_k^\textrm{T}$ are two regulator factors. Imposing the noises $\bm{\xi}^{\mathbf{W}}\mathbf{A}_k^{\textrm{T}}(\mathbf{A}_k\mathbf{A}_k^{\textrm{T}})^{-1}$ and $({\mathbf{B}_k^\textrm{T}\mathbf{B}_k})^{-1}\mathbf{B}_k^\textrm{T}\bm{\xi}^{\mathbf{W}}$ to $\mathbf{B}_k$ and $\mathbf{A}_k$ respectively can mitigate the noise amplification effect and ensure robust privacy protection.
\end{theorem}
\noindent\textit{Proof of Theorem 2}. Given the specific Gaussian distribution for the privacy budget $\varepsilon$, the noise terms $\bm{\xi}_k^{\mathbf{B}}\mathbf{A}_k\in \mathbb{R}^{m\times n} $ and $\mathbf{B}_k\bm{\xi}_k^{\mathbf{A}}\in \mathbb{R}^{m\times n} $ are expected to follow this distribution. To obtain $\bm{\xi}_k^{\mathbf{B}}$ and $\bm{\xi}_k^{\mathbf{A}}$ satisfying the condition, we solve the following least-squares problems:
\begin{equation}\label{eq18}
\begin{aligned}
&\min_{\bm{\xi}_k^{\mathbf{B}}}\|\bm{\xi}_k^{\mathbf{B}}\mathbf{A}_k-\bm{\xi}^{\mathbf{W}}\|^2; \ \ \ \min_{\bm{\xi}_k^{\mathbf{A}}}\|\mathbf{B}_k\bm{\xi}_k^{\mathbf{A}}-\bm{\xi}^{\mathbf{W}}\|^2;\\
& \textrm{s.t.}, \mathbf{B}_k, \bm{\xi}_k^{\mathbf{B}} \in \mathbb{R}^{m\times r}, \mathbf{A}_k, \bm{\xi}_k^{\mathbf{A}} \in \mathbb{R}^{r\times n}, \bm{\xi}^{\mathbf{W}}\in \mathbb{R}^{m\times n},
\end{aligned}
\end{equation}
where $\bm{\xi}^{\mathbf{W}}\in \mathbb{R}^{m\times n} $ is sampled from the Gaussian distribution. Considering that $\mathbf{B}_k$ and $\mathbf{A}_k$ are singular matrices, we can compute their pseudo-inverses via the singular value decomposition (SVD)~\cite{lange2010singular} and obtain final solutions to the above problems~\cite{peters1970least}: ${\bm{\xi}_k^{\mathbf{B}}}^\star =\bm{\xi}^{\mathbf{W}}\mathbf{A}_k^{\textrm{T}}(\mathbf{A}_k\mathbf{A}_k^{\textrm{T}})^{-1}$ and ${\bm{\xi}_k^{\mathbf{A}}}^\star=({\mathbf{B}_k^\textrm{T}\mathbf{B}_k})^{-1}\mathbf{B}_k^\textrm{T}\bm{\xi}^{\mathbf{W}}$. Here, we refer to $\mathbf{A}_k^{\textrm{T}}(\mathbf{A}_k\mathbf{A}_k^{\textrm{T}})^{-1}$ and $({\mathbf{B}_k^\textrm{T}\mathbf{B}_k})^{-1}\mathbf{B}_k^\textrm{T}$ as regulator factors. We apply ${\bm{\xi}_k^{\mathbf{B}}}^\star$ and ${\bm{\xi}_k^{\mathbf{A}}}^\star$ into Eq.~(\ref{eq171}) and Eq.~(\ref{eq172}), respectively:
\begin{equation}\label{eq19}
\begin{aligned}
\mathbf{W}_0 + \left[\mathbf{B}_k + \bm{\xi}^{\mathbf{W}}\mathbf{A}_k^{\textrm{T}}(\mathbf{A}_k\mathbf{A}_k^{\textrm{T}})^{-1}\right]\mathbf{A}_k =  \mathbf{W}_0 + \mathbf{B}_k\mathbf{A}_k + \bm{\xi}^{\mathbf{W}}, \\
\mathbf{W}_0 + \mathbf{B}_k\left[\mathbf{A}_k + ({\mathbf{B}_k^\textrm{T}\mathbf{B}_k})^{-1}\mathbf{B}_k^\textrm{T}\bm{\xi}^{\mathbf{W}}\right]=\mathbf{W}_0 + \mathbf{B}_k\mathbf{A}_k + \bm{\xi}^{\mathbf{W}},
\end{aligned}
\end{equation}
where $\mathbf{A}_k^{\textrm{T}}(\mathbf{A}_k\mathbf{A}_k^{\textrm{T}})^{-1}\mathbf{A}_k = \mathbf{B}_k({\mathbf{B}_k^\textrm{T}\mathbf{B}_k})^{-1}\mathbf{B}_k^\textrm{T}  = \mathbf{I}\in \mathbb{R}^{m \times n}$. Since $\bm{\xi}^{\mathbf{W}}$ does not undergo significant change during training, the amplification of linear noises is suppressed. \ \ \ \  \ \ \ \ \ \ \ \ \ \ \ \ \ \ \ \ \ $\square$

In Fig.~\ref{fig:noise}, we demonstrate the norms of noise terms $\|\mathbf{B}_k\bm{\xi}_k^{\mathbf{A}}\|_F$ and $\|\bm{\xi}_k^{\mathbf{B}}\mathbf{A}_k\|_F$ for our method in different communication rounds. It can be observed that the noise norms have slight fluctuations and do not present an increasing trend for any privacy budgets. Therefore, the noise amplification effect is removed with the synergy between the proposed deviation eliminator and noise regulator.

\section{Experiments}
\label{sec:experiments}
To investigate the effectiveness of the proposed DEeR, we evaluate it on two medical classification datasets (OCT-C8~\cite{subramanian2022classification} and Kvasir-v2~\cite{Pogorelov2017kvasir}) and two medical segmentation datasets (M\&MS~\cite{campello2021multi} and polyp segmentation~\cite{fan2020pranet}).
\subsection{Datasets}
\subsubsection{OCT-C8} OCT-C8~\cite{subramanian2022classification} contains $24000$ retinal OCT images, which belong to eight categories, \textit{i.e.}, age related macular degeneration, choroidal neovascularisation, diabetic macular edema, drusen, macular hole, diabetic retinopathy, central serous retinopathy and one for healthy class. Based on the official division, $18400$ images are used for training, $2800$ images for validation, and $2800$ images for testing.
\subsubsection{Kvasir-v2}  We collect $8000$ endoscopic images of the gastrointestinal tract from Kvasir-v2 dataset~\cite{Pogorelov2017kvasir}. These samples are divided into eight classes according to the types of anatomical landmarks and phatological  findings. We use the ratio of $7 : 1 : 2$ to randomly partition these samples into training, validation, and test sets.
\subsubsection{M\&MS} We gather $317$ cardiac magnetic resonance scans from different patients from M\&MS~\cite{campello2021multi}. These scans were scanned in clinical centers in three countries (Spain, Germany and Canada) using four different scanner vendors (Siemens, General Electric, Philips and Canon). Each scan is segmented into background area, left ventricular myocardium, left and right ventricle blood pools. We divide these scans into four clients based on the vendor type. The scans of each client are randomly partitioned into training, validation, and test sets with a ratio of $7 : 1 : 2$. All 3D volumes are sliced into images with the axial plane.
\subsubsection{Polyp Segmentation Dataset} The data are collected from four public datasets, CVC-ClinicDB~\cite{bernal2015wm}, CVC-ColonDB~\cite{bernal2012towards}, ETIS~\cite{silva2014toward} and Kvasir~\cite{jha2020kvasir}. Following the study~\cite{fan2020pranet}, we adopt the 900 and 550 images from ClinicDB and Kvasir datasets as the training set. The remaining 64 images of ClinicDB dataset and 100 images of Kvasir dataset belong to the test set. In addition, ETIS and CVC-ColonDB datasets contain 380 images and 196 images, respectively, which are totally divided into the test set to verify the generalization ability of a model. We randomly and evenly divided training images into four clients.

\begin{table*}[!t]
\center
\renewcommand\arraystretch{1.0}
\setlength{\tabcolsep}{2pt}
\caption{The performance comparison of different methods on two classification datasets.
}
\begin{tabular}{p{40pt}|p{45pt}|p{45pt}|p{45pt}|p{45pt}|p{45pt}|p{45pt}|p{45pt}|p{48pt}|p{48pt}}
\toprule
\multirow{2}{1.1cm}{\makecell[c]{Datasets} }&\multirow{2}{1.7cm}{\makecell[c]{Priv. Budget} }   & \multicolumn{2}{l|}{\makecell[c]{LoRA}} & \multicolumn{2}{l|}{\makecell[c]{FFA-LoRA}}& \multicolumn{2}{l|}{\makecell[c]{DP-DyLoRA}} & \multicolumn{2}{l}{\makecell[c]{DEeR}} \\
\cmidrule(r){3-10}
&                 &    \makecell[c]{Accuracy}     &   \makecell[c]{F1-score}   &   \makecell[c]{Accuracy}        &  \makecell[c]{F1-score} &   \makecell[c]{Accuracy}      &   \makecell[c]{F1-score}  &   \makecell[c]{Accuracy}      &   \makecell[c]{F1-score}   \\
\midrule
\multirow{4}{1cm}{\makecell[c]{OCT-8} }&\makecell[c]{Without DP}&\makecell[c]{$92.07\pm1.24$}  &  \makecell[c]{$92.09\pm1.23$} &  \makecell[c]{$84.86\pm2.07$} & \makecell[c]{$84.87\pm2.14$}&  \makecell[c]{$92.67\pm1.38$} & \makecell[c]{$92.66\pm1.37$}   &  \makecell[c]{$\mathbf{92.93}\pm0.82$} & \makecell[c]{$\mathbf{92.95}\pm0.82$}  \\
&\makecell[c]{$\varepsilon = 1.0$} & \makecell[c]{$80.92\pm2.42$}  & \makecell[c]{$79.77\pm3.11$}  &  \makecell[c]{$76.91\pm4.15$} & \makecell[c]{$76.65\pm4.17$}&  \makecell[c]{$79.26\pm3.89$} & \makecell[c]{$77.72\pm5.07$}  & \makecell[c]{$\mathbf{92.33}\pm0.55$}  & \makecell[c]{$\mathbf{92.35}\pm0.55$}  \\
&\makecell[c]{$\varepsilon = 0.5$}& \makecell[c]{$83.39\pm3.88$}  &\makecell[c]{$82.30\pm5.03$} & \makecell[c]{$74.53\pm4.39$} & \makecell[c]{$72.57\pm5.54$}&  \makecell[c]{$61.70\pm1.89$} & \makecell[c]{$58.36\pm4.24$} & \makecell[c]{$\mathbf{91.20}\pm0.95$}  & \makecell[c]{$\mathbf{91.21}\pm0.97$}  \\
&\makecell[c]{$\varepsilon = 0.1$}&\makecell[c]{$42.15\pm2.88$}  &\makecell[c]{$38.59\pm3.60$}  &\makecell[c]{$56.20\pm4.63$} & \makecell[c]{$52.29\pm6.10$}&  \makecell[c]{$23.55\pm4.61$} & \makecell[c]{$19.30\pm3.40$} & \makecell[c]{$\mathbf{84.28}\pm3.74$} & \makecell[c]{$\mathbf{83.20}\pm4.59$} \\
\midrule
\multirow{4}{1cm}{\makecell[c]{Kvasir-v2} }&\makecell[c]{Without DP}&  \makecell[c]{$85.46\pm1.39$}  & \makecell[c]{$85.18\pm2.00$ }    & \makecell[c]{$80.73\pm1.08$}  &  \makecell[c]{$79.90\pm1.56$}&  \makecell[c]{$76.10\pm5.59$}  &  \makecell[c]{$74.15\pm6.92$}  &  \makecell[c]{$\mathbf{86.29}\pm0.65$}  & \makecell[c]{$\mathbf{86.11}\pm0.65$}    \\
&\makecell[c]{$\varepsilon = 6.0$} & \makecell[c]{$84.58\pm0.57$}  &\makecell[c]{$84.25\pm0.58$}  &\makecell[c]{$79.23\pm0.87$} & \makecell[c]{$78.14\pm1.03$}&  \makecell[c]{$78.85\pm3.92$} & \makecell[c]{$77.57\pm5.50$} & \makecell[c]{$\mathbf{87.00}\pm0.87$} &\makecell[c]{$\mathbf{86.84}\pm0.91$}  \\
&\makecell[c]{$\varepsilon = 3.0$}& \makecell[c]{$84.85\pm0.86$}  & \makecell[c]{$84.51\pm0.86$} & \makecell[c]{$79.21\pm0.62$}  & \makecell[c]{$77.94\pm0.80$}&  \makecell[c]{$76.93\pm1.40$}  & \makecell[c]{$75.17\pm2.68$} & \makecell[c]{$\mathbf{86.90}\pm0.75$} &\makecell[c]{$\mathbf{86.70}\pm0.85$}  \\
&\makecell[c]{$\varepsilon = 1.0$}& \makecell[c]{$82.00\pm0.82$}  & \makecell[c]{$81.01\pm1.11$} & \makecell[c]{$79.06\pm1.35$}  &\makecell[c]{$77.90\pm1.51$}&  \makecell[c]{$76.37\pm 2.45$}  &\makecell[c]{$73.69\pm3.86$} & \makecell[c]{$\mathbf{86.56}\pm0.53$} &\makecell[c]{$\mathbf{86.33}\pm0.63$} \\
 \bottomrule
\end{tabular}
\label{tab:classification}
\vspace{-2.0mm}
\end{table*}

\begin{table*}[!t]
\center
\renewcommand\arraystretch{1.0}
\setlength{\tabcolsep}{2pt}
\caption{The performance comparison of different methods on cardiac image segmentation dataset. 
}
\begin{tabular}{p{40pt}|p{45pt}|p{45pt}|p{45pt}|p{45pt}|p{45pt}|p{45pt}|p{45pt}|p{48pt}|p{48pt}}
\toprule
\multirow{2}{1.1cm}{\makecell[c]{Priv. Budget} }& \multirow{2}{1.5cm}{\centering\makecell[c]{Clients} }   & \multicolumn{2}{l|}{\makecell[c]{LoRA}} & \multicolumn{2}{l|}{\makecell[c]{FFA-LoRA}}& \multicolumn{2}{l|}{\makecell[c]{DP-DyLoRA}} & \multicolumn{2}{l}{\makecell[c]{DEeR}} \\
\cmidrule(r){3-10}
&                 &    \makecell[c]{IoU}     &   \makecell[c]{Dice}   &   \makecell[c]{IoU}        &  \makecell[c]{Dice} &   \makecell[c]{IoU}      &   \makecell[c]{Dice} &   \makecell[c]{IoU}      &   \makecell[c]{Dice}  \\
\midrule
\multirow{4}{0cm}{\makecell[c]{$\varepsilon = 1.0$} }&\makecell[c]{Canon}& \makecell[c]{$73.73\pm1.77$}  & \makecell[c]{$82.99\pm1.57$} & \makecell[c]{$76.00\pm0.78$}  &\makecell[c]{$84.72\pm0.80$} & \makecell[c]{$74.87\pm1.43$} & \makecell[c]{$84.19\pm1.40$}  & \makecell[c]{$\mathbf{77.22}\pm1.32$} &\makecell[c]{$\mathbf{85.63}\pm1.21$}   \\
&\makecell[c]{GE} & \makecell[c]{$73.32\pm1.59$}  & \makecell[c]{$82.89\pm1.61$} & \makecell[c]{$74.98\pm0.85$}  &\makecell[c]{$83.83\pm0.76$}&  \makecell[c]{$73.90\pm1.06$} & \makecell[c]{$83.52\pm0.98$}  & \makecell[c]{$\mathbf{76.37}\pm1.39$} &\makecell[c]{$\mathbf{84.97}\pm1.33$}  \\
&\makecell[c]{Philips}& \makecell[c]{$75.54\pm0.34$}  & \makecell[c]{$84.83\pm0.36$} & \makecell[c]{$77.28\pm0.69$}  &\makecell[c]{$86.03\pm0.61$}& \makecell[c]{$74.40\pm0.18$} & \makecell[c]{$84.02\pm0.13$}   & \makecell[c]{$\mathbf{78.83}\pm0.04$} &\makecell[c]{$\mathbf{87.31}\pm0.11$}  \\
&\makecell[c]{Siemens}&\makecell[c]{$75.15\pm1.03$}  & \makecell[c]{$83.91\pm1.06$} & \makecell[c]{$77.26\pm1.19$}  &\makecell[c]{$85.47\pm1.19$}& \makecell[c]{$73.71\pm0.58$} & \makecell[c]{$82.96\pm0.24$}  & \makecell[c]{$\mathbf{78.45}\pm1.06$} &\makecell[c]{$\mathbf{86.51}\pm1.03$}  \\
\midrule
\multirow{4}{0cm}{\makecell[c]{$\varepsilon = 0.1$} }&\makecell[c]{Canon}&  \makecell[c]{$69.65\pm0.77$}  & \makecell[c]{$79.94\pm0.57$} & \makecell[c]{$76.41\pm1.56$}  &\makecell[c]{$84.96\pm1.48$}& \makecell[c]{$70.02\pm1.31$} & \makecell[c]{$80.37\pm1.21$}   & \makecell[c]{$\mathbf{77.52}\pm1.68$} &\makecell[c]{$\mathbf{85.87}\pm1.49$}   \\
&\makecell[c]{GE} & \makecell[c]{$69.18\pm1.90$}  & \makecell[c]{$79.82\pm1.69$} & \makecell[c]{$74.83\pm1.81$}  &\makecell[c]{$83.78\pm1.70$}& \makecell[c]{$70.08\pm1.25$} & \makecell[c]{$80.73\pm1.31$}  & \makecell[c]{$\mathbf{75.37}\pm1.72$} &\makecell[c]{$\mathbf{84.33}\pm1.41$} \\
&\makecell[c]{Philips}&  \makecell[c]{$ 70.62\pm0.81$}  & \makecell[c]{$81.04\pm0.56$} & \makecell[c]{$77.02\pm0.23$}  &\makecell[c]{$85.94\pm0.17$}& \makecell[c]{$72.22\pm0.58$} & \makecell[c]{$82.10\pm0.42$}  & \makecell[c]{$\mathbf{78.02}\pm0.25$} &\makecell[c]{$\mathbf{86.65}\pm0.18$} \\
&\makecell[c]{Siemens}& \makecell[c]{$69.75\pm1.25$}  & \makecell[c]{$79.97\pm1.36$} & \makecell[c]{$76.72\pm0.73$}  &\makecell[c]{$85.11\pm0.89$}& \makecell[c]{$71.03\pm1.48$} & \makecell[c]{$80.96\pm1.20$}  & \makecell[c]{$\mathbf{78.18}\pm0.84$} &\makecell[c]{$\mathbf{86.43}\pm0.78$} \\
 \bottomrule
\end{tabular}
\label{tab:cardiac}
\end{table*}

\subsection{Experiment Setup}
\subsubsection{Implementation Details}
The proposed DEeR and comparison methods are implemented with PyTorch library. For classification datasets, BiomedCLIP~\cite{zhang2023large} is regarded as the foundation model. The number $K$ of clients is set to $12$. We keep the total communication rounds to $50$ and the local steps to $5$. The total batch-sizes are set to $128$ and $512$ for Kvasir-v2 and OCT-C8, respectively. We use Dirichlet distribution on label ratios to simulate Non-IID settings. The Dirichlet parameter $\beta$ defaults to $0.1$. For segmentation datasets, we use SAM-Med2D~\cite{cheng2023sam} as the foundation model and box as prompt. We keep the total communication rounds to $50$ and the local steps to $3$. The total batch-size is set to $32$ for M\&MS dataset and $128$ for polyp segmentation dataset. For all datasets, we use the SGD optimizer and choose the best learning rate from $[0.1, 0.01, 0.001]$ by FedAvg with LoRA. Both the rank $r$ and scaling factor $\alpha$ default to 8. For privacy parameters, the privacy failure probability $\delta=\frac{1}{K}$. The privacy budget $\varepsilon$ defaults to $3$ for Kvasir-v2 and 0.1 for OCT-8 and M\&MS. We use the privacy accountant from Opacus~\cite{opacus} to calculate the noise scale $\sigma$ in all experiments. The clipping threshold $C$ is selected by grid search from set $[0.1, 0.2, 0.3, 0.4, 0.6]$.
\subsubsection{Evaluation Metrics}
Two commonly-used metrics, accuracy, and F1-score, are used to measure the classification performance. To evaluate segmentation performance,  we adopt two commonly-used metrics of Dice similarity coefficient and mean intersection over union (IoU) of foreground and background. In all the experiments, we conduct three trials for each setting and present the mean and the standard deviation.

\subsection{Comparisons with State-of-the-Art Methods}
We compare DEeR with three baselines on different medical tasks. (1) LoRA: We implement the original LoRA~\cite{hu2021lora} based on FedAvg~\cite{mcmahan2017communication}.  (2) FFA-LoRA~\cite{sun2024improving}: It freezes $\mathbf{A}$ and only finetunes $\mathbf{B}$ of LoRA in FedAvg. (3) DP-DyLoRA~\cite{xu2024dp}: it adjusts the rank $r$ of LoRA layers randomly during training, in the range of $r \in [r_{\textrm{min}}, r_{\textrm{max}}]$. During testing, we reported the best results among these ranks.

\subsubsection{Evaluation on Medical Classification Tasks} To verify efficiency of DEeR for medical classification tasks,  we compare performance of DEeR and baseline methods, under different privacy budgets $\varepsilon \in [1.0, 3.0, 6.0]$ for Kvasir-v2 dataset and $\varepsilon \in [0.1, 0.5, 1.0]$ for OCT-8 dataset, as shown in Table~\ref{tab:classification}.
For Kvasir-v2 dataset, LoRA yields the second-best performance and undergoes a severe performance degradation as $\varepsilon$ becomes smaller, especially F1-score with a decrement of $4.17\%$. Although FFA-LoRA presents relatively stable performance against varying privacy budgets, it obtains the lowest accuracy and F1-score. By comparison, DEeR implements the highest accuracy and F1-score and also shows consistent performance for different privacy budgets. In experiments, we also observe that recall scores of all cases in DEeR are higher than 80\%. Significant difference is found in LoRA and DEeR for $\varepsilon=1.0$ ($P$-value $<0.005$) and $\varepsilon=6.0$ ($P$-value $<0.05$).
For OCT-8 dataset, both LoRA and FFA-LoRA present the high sensitivity to $\varepsilon$. For instance, when the budget $\varepsilon$ decreases from infinity (without DP) to $0.1$, they have enormous performance drops with decrements of $49.92\%$ and $53.5\%$, $28.65\%$ and $32.58\%$ in accuracy and F1-score, respectively. Noticeably, DEeR merely suffers from a slight drop ($8.65\%$) and ($9.75\%$) and outperforms LoRA and FFA-LoRA ($P$-value $<0.0005$ for $\varepsilon=0.1$ and $P$-value $<0.005$ for $\varepsilon=1.0$). DP-DyLoRA can perform well without DP noise, but performs poorly once imposing noise and also presents a higher sensitivity to $\varepsilon$ than DEeR.  Although DP-DyLoRA presents relatively stable performance as DEeR against varying privacy budgets, it obtains low accuracy and F1-score. The results on two datasets prove that DEeR can achieve superior finetuning performance while providing stronger privacy guarantees than existing methods for medical classification tasks.

\begin{figure}[!t]	
	\begin{center}
		\subfigure[LoRA]
		{\includegraphics[width=0.49\columnwidth]{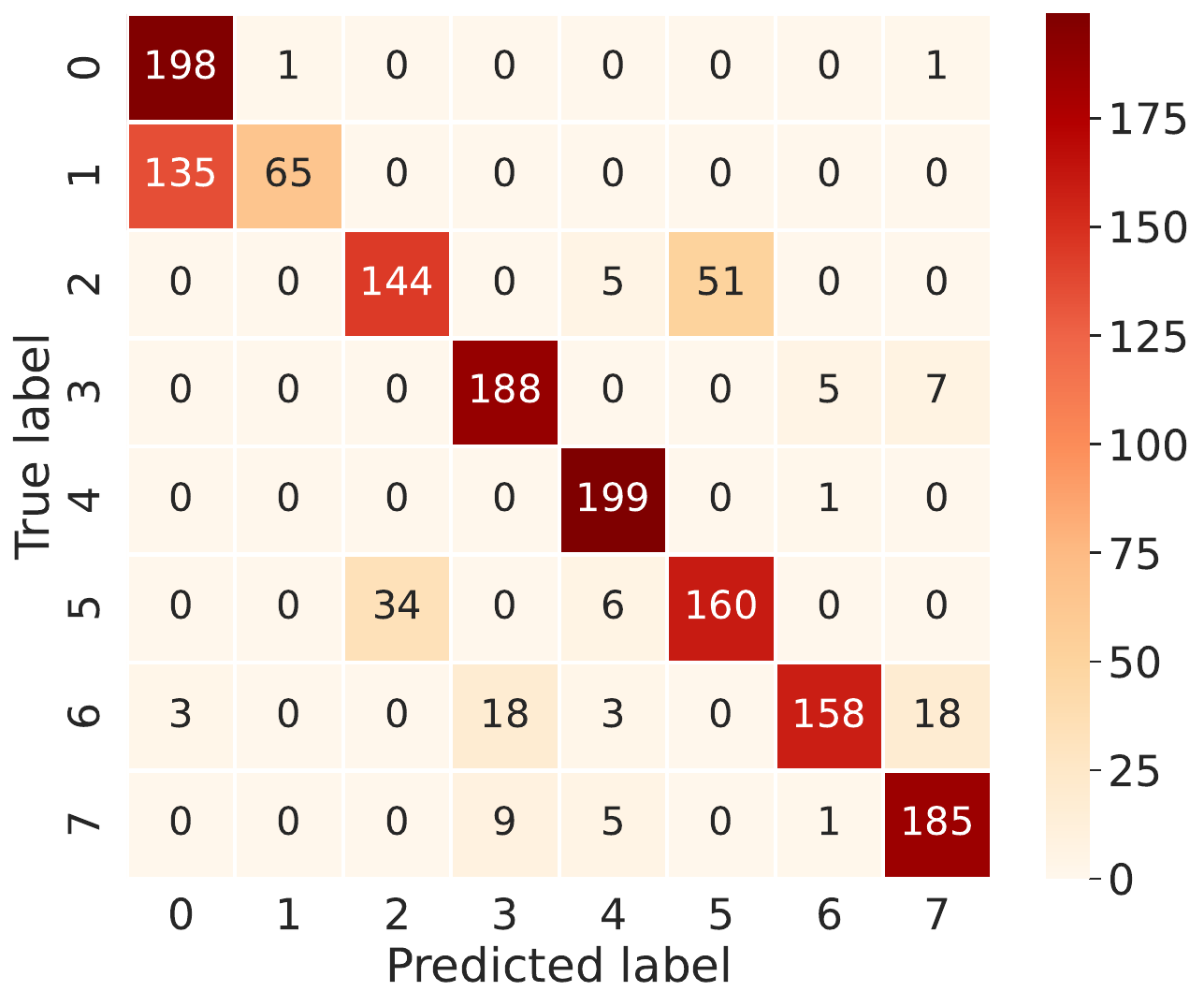}\label{fig:LoRA}}
		\subfigure[FFA-LoRA]
		{\includegraphics[width=0.49\columnwidth]{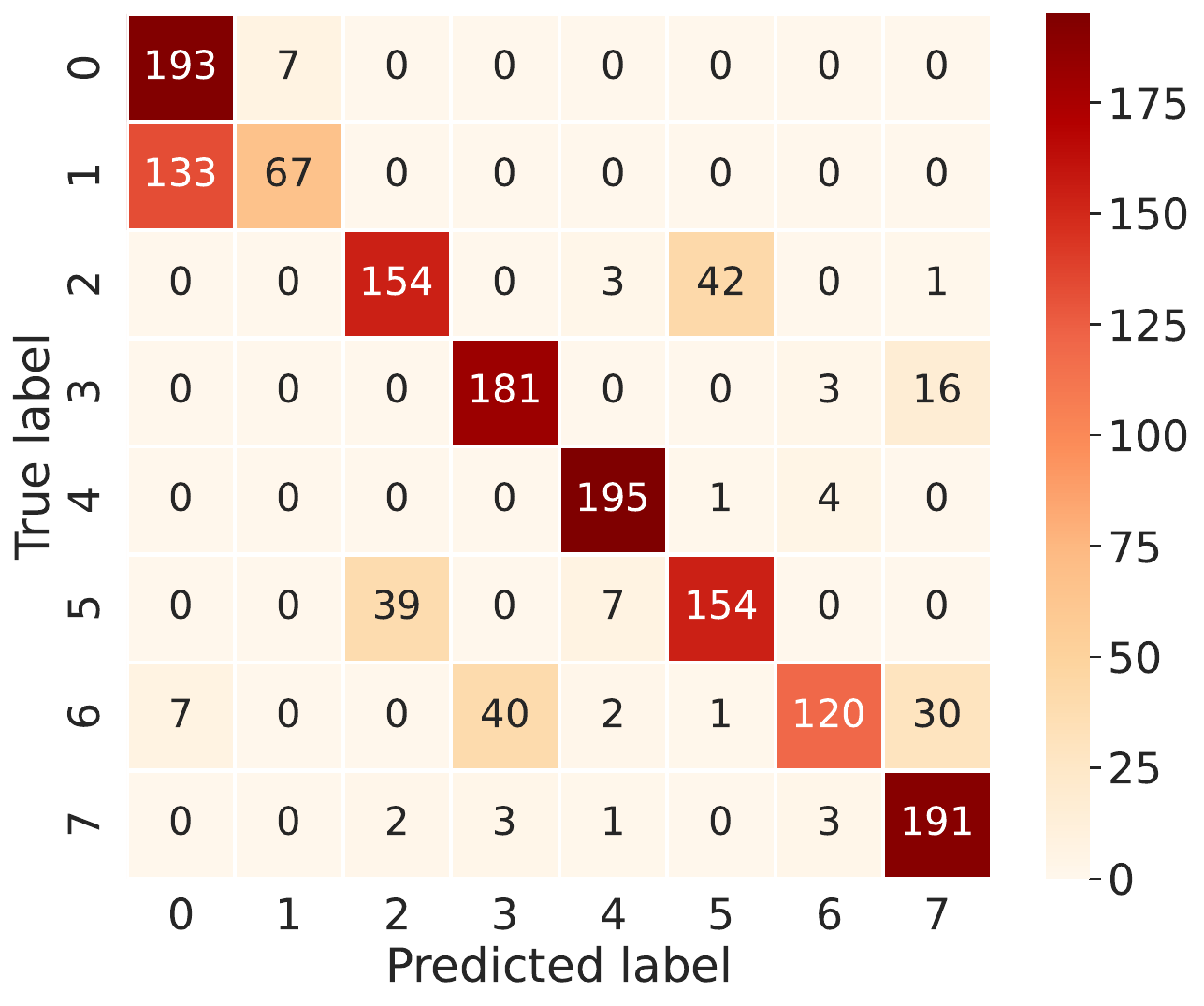}\label{fig:FFA_LoRA}}
		\subfigure[DP-DyLoRA]
		{\includegraphics[width=0.49\columnwidth]{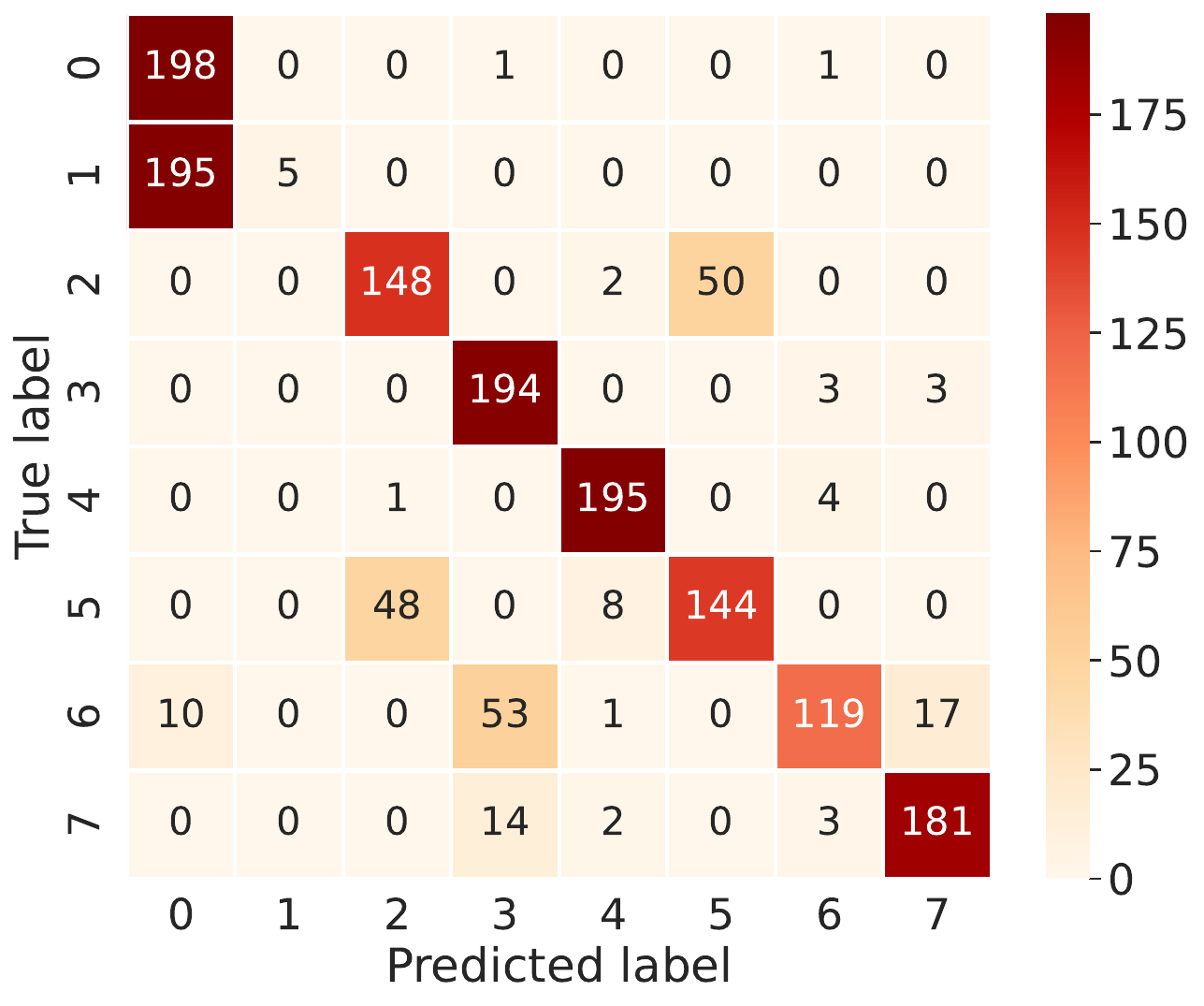}\label{fig:DP_DyLoRA}}
		\subfigure[Ours]
		{\includegraphics[width=0.49\columnwidth]{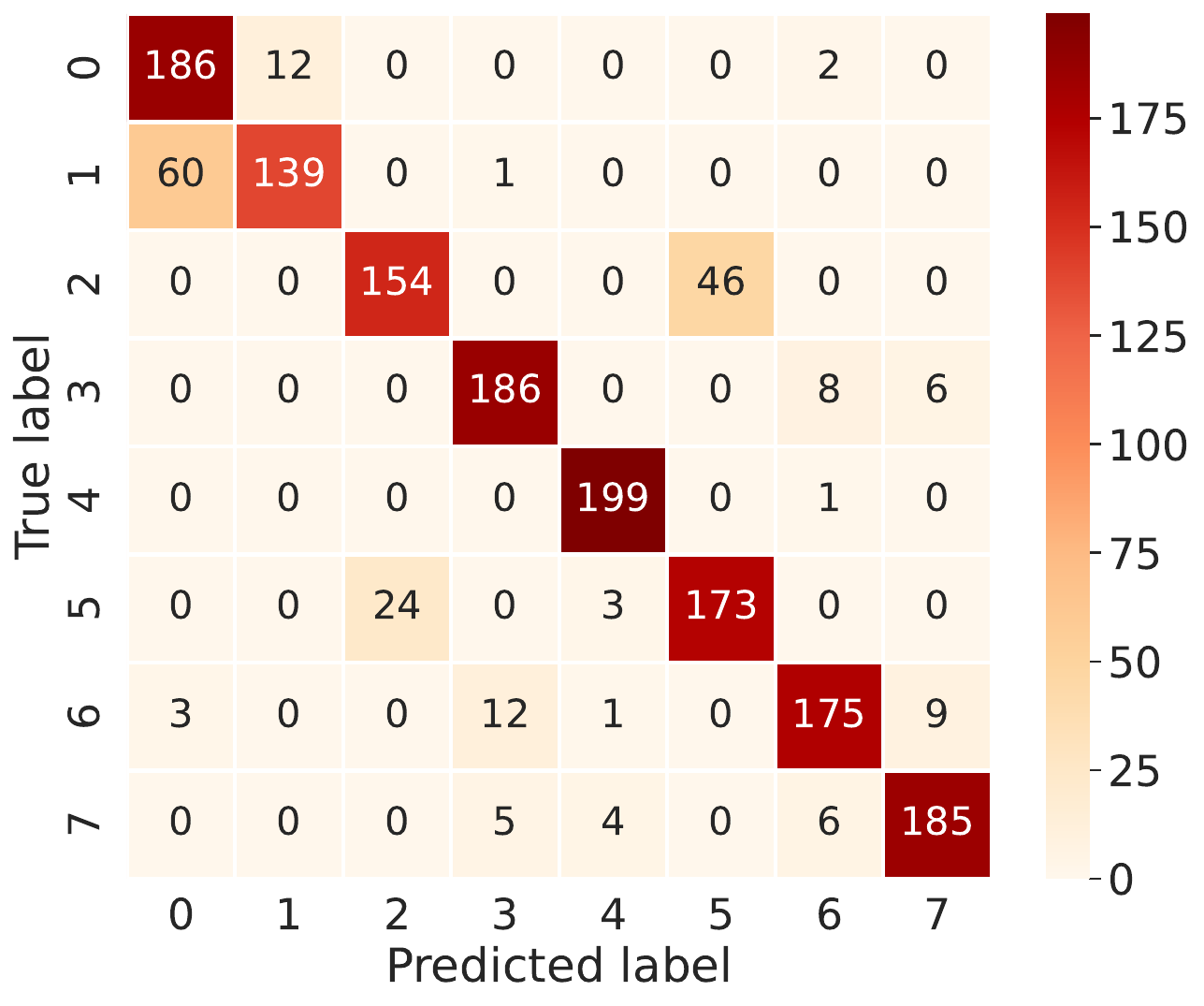}\label{fig:Ours}}
	\end{center}
	\vspace{-3mm}
	\caption{The confusion matrices of different methods on Kvasir-v2.}
\label{fig:conf_matrices}
	\vspace{-3mm}
\end{figure}

We further visualize the confusion matrices of the previous methods and our DEeR on the endoscopy dataset, as shown in Fig.~\ref{fig:conf_matrices}. We can observe that DP-DyLoRA, LoRA and FFA-LoRA misclassify 97.5\%, 67.5\% and 66.5\% the class 1 into the class 0 due to  the narrow intra-class distance, respectively. Meanwhile, they also have high errors for the class 5 and 6. By comparison, DEeR achieves higher precision in these classes, especially for the class 1. The experimental results can confirm the effectiveness of the proposed finetuning method.
\begin{table}[!t]
\center
\renewcommand\arraystretch{1.1}
\setlength{\tabcolsep}{0.5pt}
\caption{The performance comparison of different methods on polyp segmentation dataset under different privacy budgets.}
\begin{tabular}{p{43pt}|p{50pt}|p{50pt}|p{50pt}|p{50pt}}
\toprule
\multirow{2}{1.5cm}{\makecell[c]{Methods} } & \multicolumn{2}{l|}{\makecell[c]{$\varepsilon = 1.0$}} & \multicolumn{2}{l}{\makecell[c]{$\varepsilon = 0.1$}} \\
\cmidrule(r){2-5}
                  & \makecell[c]{IoU (\%)}& \makecell[c]{Dice (\%)} & \makecell[c]{IoU (\%)}& \makecell[c]{Dice (\%)}           \\
\midrule
\makecell[c]{LoRA} & \makecell[c]{$78.83\pm0.08$}   & \makecell[c]{$87.17\pm0.03$}   & \makecell[c]{$75.98\pm0.62$}   &\makecell[c]{$85.44\pm0.31$}  \\
\makecell[c]{FFA-LoRA}&\makecell[c]{$80.96\pm0.04$}  & \makecell[c]{$88.60\pm0.03$}  & \makecell[c]{$79.91\pm0.18$}  &\makecell[c]{$87.84\pm0.07$}  \\
\makecell[c]{DP-DyLoRA} &\makecell[c]{$73.84\pm1.00$}&\makecell[c]{$83.90\pm0.43$} & \makecell[c]{$75.00\pm1.01$} & \makecell[c]{$84.69\pm0.43$}  \\
\makecell[c]{DEeR} & \makecell[c]{$\textbf{81.50}\pm0.22$}  &\makecell[c]{$\textbf{88.92}\pm0.13$}  &\makecell[c]{$\textbf{80.60}\pm0.15$}  &\makecell[c]{$\textbf{88.38}\pm0.04$}  \\
\bottomrule
\end{tabular}
\label{tab:polyp}
\vspace{-3mm}
\end{table}
\subsubsection{Evaluation on Medical Segmentation Tasks} We compare DEeR with baseline methods on M\&MS and polyp segmentation datasets with privacy budgets $\varepsilon \in [0.1, 1.0]$. For M\&MS in Table~\ref{tab:cardiac}, it is observed that the performance of LoRA is fragile for the budget $\varepsilon$, since its IoU and Dice on all clients suffer from remarkable decreases when $\varepsilon$ declines from $1$ to $0.1$. Different from classification tasks, FFA-LoRA outperforms LoRA on segmentation tasks. One possible reason is that the segmentation model is more sensitive to noise. Compared with LoRA, FFA-LoRA is not affected by ``quadratic'' noise. Nonetheless, it is still inferior to DEeR in terms of performance for different privacy budgets $\varepsilon$ since it neglects the impact of ``linear'' noise. The comparison results of polyp segmentation are demonstrated in Table~\ref{tab:polyp}. LoRA and DP-DyLoRA~\cite{xu2024dp} show the limited performance for different privacy budget $\varepsilon$, since they ignore aggregation deviation and noise amplification effect problems. In contrast, FFA-LoRA exploits a simple freezing strategy to address these problems and achieves the better performance. Notably, DEeR outperforms FFA-LoRA for any $\varepsilon$ and yields approximating 90\% of dice scores. These results confirm the priority of DEeR for medical segmentation tasks in contrast to the state-of-the-art methods.
\begin{table*}[!t]
\center
\renewcommand\arraystretch{1.0}
\setlength{\tabcolsep}{4pt}
\caption{The performance of the proposed federated finetuning framework with different modules. 
}
\begin{tabular}{p{40pt}|p{45pt}|p{50pt}|p{50pt}|p{50pt}|p{50pt}|p{50pt}|p{50pt}}
\toprule
\multirow{2}{1.1cm}{\makecell[c]{Datasets} }&\multirow{2}{1.7cm}{\makecell[c]{Priv. Budget} }      & \multicolumn{2}{l|}{\makecell[c]{w/o Deviation Eliminator }} & \multicolumn{2}{l|}{\makecell[c]{w/o Noise Regulator}} & \multicolumn{2}{l}{\makecell[c]{DEeR}} \\
\cmidrule(r){3-8}
&                 &    \makecell[c]{Accuracy (\%)}     &   \makecell[c]{F1-score (\%)}   &   \makecell[c]{Accuracy (\%)}        &  \makecell[c]{F1-score (\%)} &   \makecell[c]{Accuracy (\%)}      &   \makecell[c]{F1-score (\%)}   \\
\midrule
\multirow{4}{0cm}{\makecell[c]{OCT-8} }&\makecell[c]{$\varepsilon = 1.0$}&  \makecell[c]{$43.83\pm5.05$}  & \makecell[c]{$35.18\pm4.47$} & \makecell[c]{$86.25\pm3.77$} & \makecell[c]{$85.78\pm4.25$} & \makecell[c]{$\mathbf{92.33}\pm0.55$}  & \makecell[c]{$\mathbf{92.35}\pm0.55$}  \\
&\makecell[c]{$\varepsilon = 0.5$} &  \makecell[c]{$34.44\pm7.65$}  & \makecell[c]{$25.37\pm6.80$} & \makecell[c]{$82.61\pm5.43$}  & \makecell[c]{$81.53\pm6.26$} & \makecell[c]{$\mathbf{91.20}\pm0.95$}  & \makecell[c]{$\mathbf{91.21}\pm0.97$}  \\
&\makecell[c]{$\varepsilon = 0.1$}& \makecell[c]{$15.89\pm3.95$}  & \makecell[c]{$10.25\pm4.64$} & \makecell[c]{$69.44\pm4.36$}  & \makecell[c]{$67.45\pm6.14$} & \makecell[c]{$\mathbf{84.28}\pm3.74$} & \makecell[c]{$\mathbf{83.20}\pm4.59$} \\
\midrule
\multirow{4}{0cm}{\makecell[c]{Kvasir-v2} }&\makecell[c]{$\varepsilon = 6.0$}&  \makecell[c]{$72.43\pm5.29$}  & \makecell[c]{$68.93\pm5.31$ }    & \makecell[c]{$85.77\pm0.43$}  &  \makecell[c]{$85.46\pm 0.35$}  &  \makecell[c]{$\mathbf{87.00}\pm0.87$}  & \makecell[c]{$\mathbf{86.84}\pm0.91$}    \\
&\makecell[c]{$\varepsilon = 3.0$} & \makecell[c]{$72.16\pm3.03$}  &\makecell[c]{$67.91\pm2.62$}  &\makecell[c]{$84.81\pm0.30$} & \makecell[c]{$84.49\pm0.48$} & \makecell[c]{$\mathbf{86.90}\pm0.75$} &\makecell[c]{$\mathbf{86.70}\pm0.85$}  \\
&\makecell[c]{$\varepsilon = 1.0$}& \makecell[c]{$65.93\pm1.00$}  & \makecell[c]{$60.98\pm1.95$} & \makecell[c]{$78.56\pm1.68$}  & \makecell[c]{$77.01\pm2.46$} & \makecell[c]{$\mathbf{86.56}\pm0.53$} &\makecell[c]{$\mathbf{86.33}\pm0.63$}  \\
 \bottomrule
\end{tabular}
\label{tab:ablation}
\end{table*}

\begin{figure}[!t]
\centering
\vspace{2.0mm}
\includegraphics[width = 0.98\columnwidth]{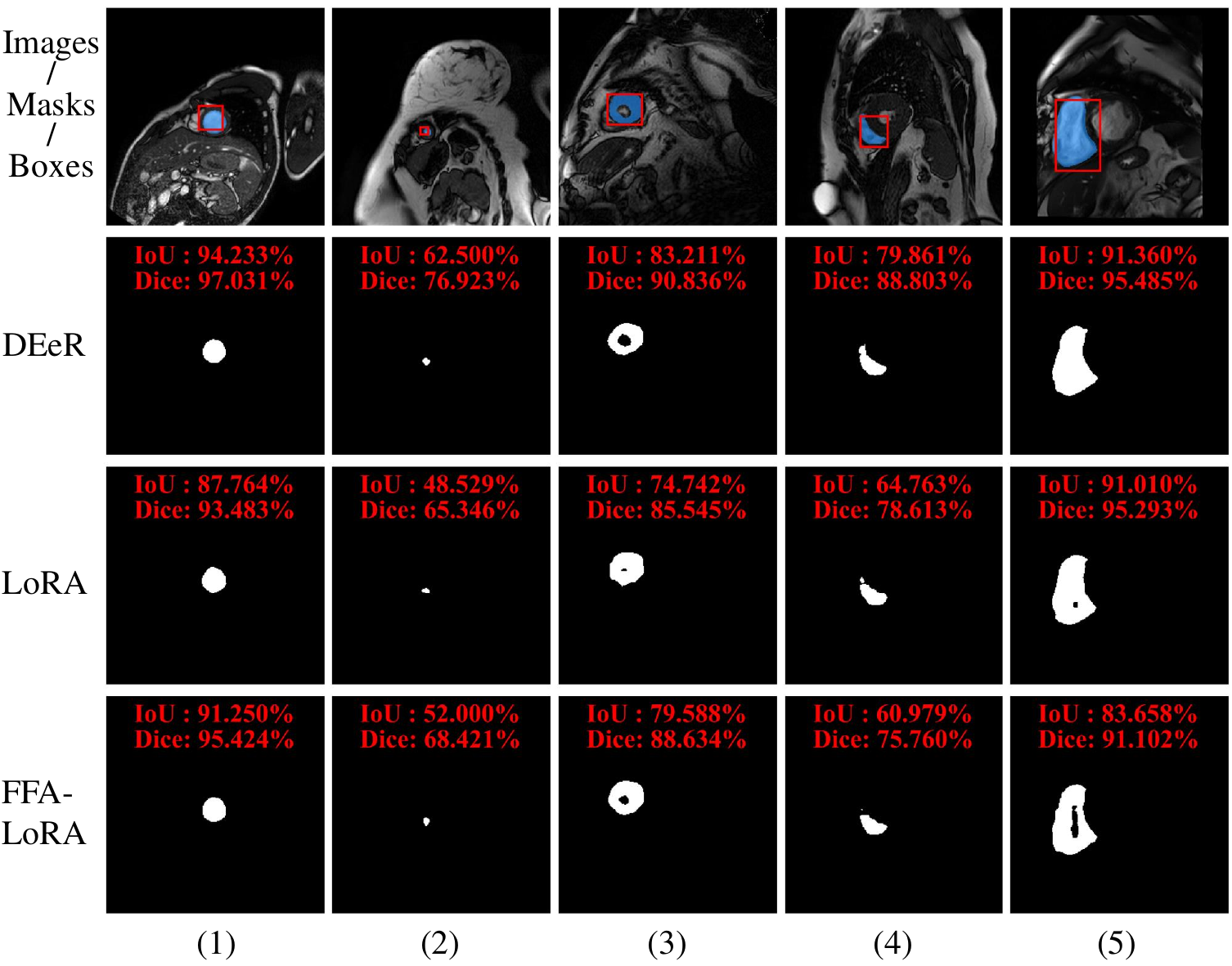}
\caption{ Visualization of segmentation results for different methods on M\&MS dataset. The columns (1)-(3) correspond to $\varepsilon = 0.1$ and columns (4)-(5) correspond to $\varepsilon = 1.0$. }
\label{fig:seg_results_cardiac}
\vspace{-2.0mm}
\end{figure}

\begin{figure}[!t]
\centering
\vspace{2.0mm}
\includegraphics[width = 0.98\columnwidth]{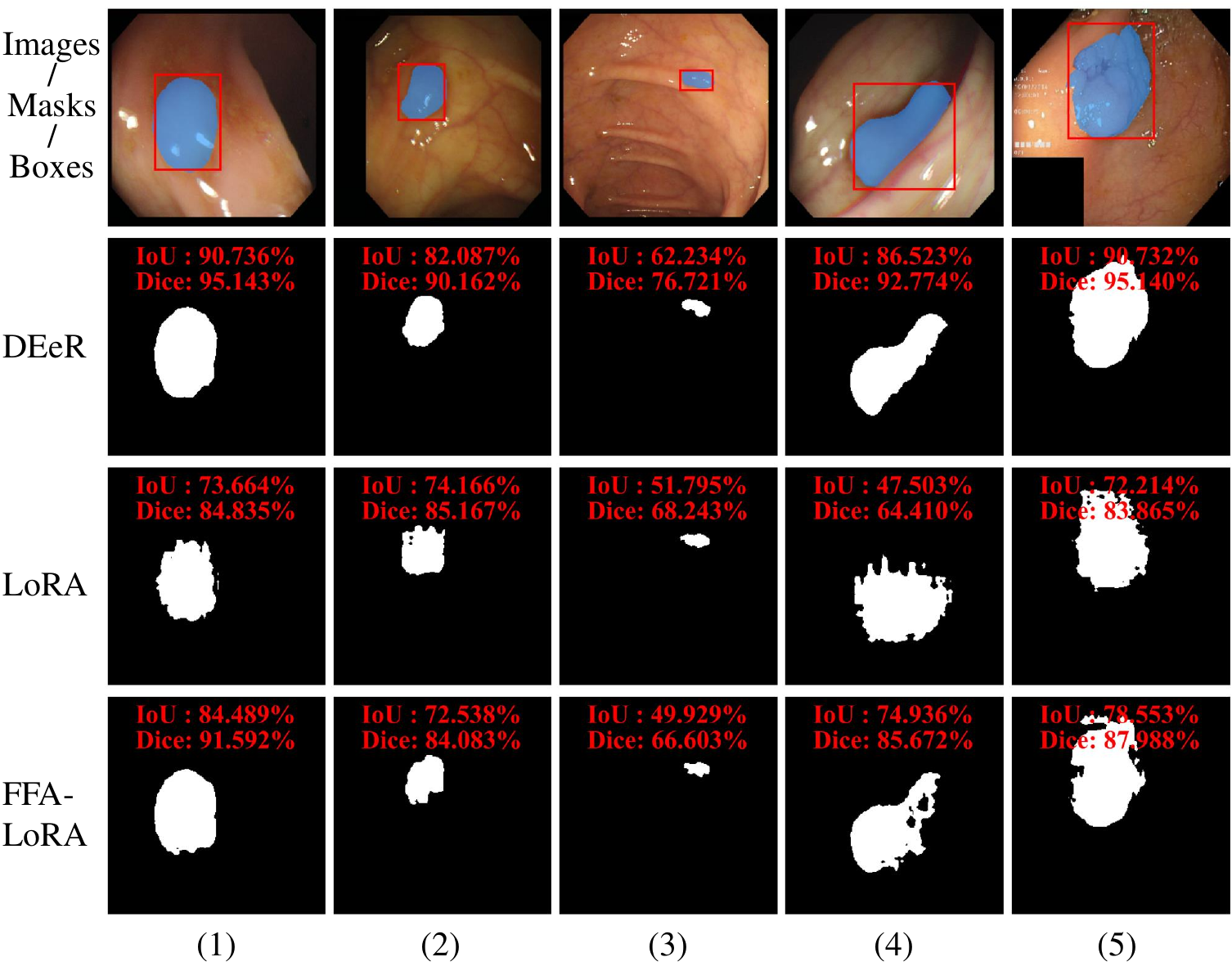}
\caption{ Visualization of segmentation results for different methods on polyp segmentation dataset.}
\label{fig:seg_results_polyp}
\vspace{-2.0mm}
\end{figure}

Furthermore, we visualize the segmentation results of DEeR and the state-of-the-art methods under different privacy budgets $\varepsilon$, as shown in Fig.~\ref{fig:seg_results_cardiac}. The simple case of the 1-\textit{st} column is accurately segmented by all methods. Noticeably, our method obtains higher performance since it has a superior capacity to detect boundaries. Although some object regions are small (2-\textit{nd} column), discontinuous (3-\textit{rd} column),  or irregular (4-\textit{th} and 5-\textit{th} columns),  DEeR can also more accurately segment them than LoRA and FFA-LoRA. We also visualize the polyp segmentation results of DEeR
and previous methods, as shown in Fig.~\ref{fig:seg_results_polyp}. We can find that DEeR can more accurately segment various polyps
than LoRA and FFA-LoRA. These qualitative results further illustrate the effectiveness of our DEeR.

\begin{table*}[!t]
\center
\renewcommand\arraystretch{1.0}
\setlength{\tabcolsep}{2pt}
\caption{The performance comparison of different methods under different data heterogeneity. 
}
\begin{tabular}{p{40pt}|p{45pt}|p{45pt}|p{45pt}|p{45pt}|p{45pt}|p{45pt}|p{46pt}|p{48pt}|p{48pt}}
\toprule
\multirow{2}{1.1cm}{\makecell[c]{Datasets} }&\multirow{2}{1.5cm}{\makecell[c]{Heterogeneity} } & \multicolumn{2}{l|}{\makecell[c]{LoRA}} & \multicolumn{2}{l|}{\makecell[c]{FFA-LoRA}}& \multicolumn{2}{l|}{\makecell[c]{DP-DyLoRA}} & \multicolumn{2}{l}{\makecell[c]{DEeR}} \\
\cmidrule(r){3-10}
&                 &    \makecell[c]{Accuracy}     &   \makecell[c]{F1-score }   &   \makecell[c]{Accuracy}        &  \makecell[c]{F1-score} &   \makecell[c]{Accuracy}      &   \makecell[c]{F1-score} &   \makecell[c]{Accuracy}      &   \makecell[c]{F1-score}   \\
\midrule
\multirow{4}{1cm}{\makecell[c]{OCT-8} }&\makecell[c]{$\beta=10.0$}&\makecell[c]{$71.16\pm1.58$}  &  \makecell[c]{$70.69\pm2.11$} &  \makecell[c]{$86.21\pm0.25$} & \makecell[c]{$86.19\pm0.29$}  &  \makecell[c]{$33.94\pm2.32$} & \makecell[c]{$29.83\pm2.16$}    &  \makecell[c]{$\mathbf{94.85}\pm0.08$}  & \makecell[c]{$\mathbf{94.86}\pm0.08$}  \\
&\makecell[c]{$\beta=1.0$} & \makecell[c]{$69.51\pm0.16$}  & \makecell[c]{$68.48\pm0.71$}  & \makecell[c]{$83.27\pm1.65$}  & \makecell[c]{$83.17\pm1.62$}&  \makecell[c]{$32.86\pm1.85$} & \makecell[c]{$28.25\pm2.27$}  & \makecell[c]{$\mathbf{94.82}\pm0.28$}  & \makecell[c]{$\mathbf{94.81}\pm0.28$}  \\
&\makecell[c]{$\beta=0.5$}& \makecell[c]{$66.67\pm1.76$}  &\makecell[c]{$66.41\pm1.80$} & \makecell[c]{$81.29\pm2.17$} & \makecell[c]{$81.22\pm2.11$}&  \makecell[c]{$32.76\pm2.30$} & \makecell[c]{$30.28\pm3.00$} & \makecell[c]{$\mathbf{94.16}\pm0.28$}  & \makecell[c]{$\mathbf{94.16}\pm0.28$}  \\
&\makecell[c]{$\beta=0.1$}&\makecell[c]{$42.15\pm2.88$}  &\makecell[c]{$38.59\pm3.60$}  & \makecell[c]{$56.20\pm4.63$} & \makecell[c]{$52.29\pm6.10$}&  \makecell[c]{$23.55\pm4.61$} & \makecell[c]{$19.30\pm3.40$}   & \makecell[c]{$\mathbf{84.28}\pm3.74$} & \makecell[c]{$\mathbf{83.20}\pm4.59$} \\
\midrule
\multirow{4}{1cm}{\makecell[c]{Kvasir-v2} }&\makecell[c]{$\beta=10.0$}&\makecell[c]{$91.00\pm0.33$}  &  \makecell[c]{$90.99\pm0.34$} &  \makecell[c]{$86.54\pm1.18$} & \makecell[c]{$86.52\pm1.17$}&   \makecell[c]{$88.54\pm0.51$}  &  \makecell[c]{$88.50\pm0.54$}    &  \makecell[c]{$\mathbf{91.48}\pm0.52$} & \makecell[c]{$\mathbf{91.47}\pm0.51$}  \\
&\makecell[c]{$\beta=1.0$}&\makecell[c]{$89.98\pm0.46$} & \makecell[c]{$89.95\pm0.46$} & \makecell[c]{$86.06\pm0.44$} & \makecell[c]{$86.04\pm0.44$}& \makecell[c]{$88.29\pm0.26$} & \makecell[c]{$88.29\pm0.23$} & \makecell[c]{$\mathbf{90.38}\pm0.20$}  &\makecell[c]{$\mathbf{90.37}\pm0.21$} \\
&\makecell[c]{$\beta=0.5$}& \makecell[c]{$89.00\pm0.67$}  &\makecell[c]{$88.94\pm 0.70$} & \makecell[c]{$84.54\pm0.51$} & \makecell[c]{$84.45\pm0.54$}&  \makecell[c]{$86.60\pm2.02$}  & \makecell[c]{$86.35\pm2.27$} & \makecell[c]{$\mathbf{90.27}\pm0.46$}  & \makecell[c]{$\mathbf{90.25}\pm0.49$}  \\
&\makecell[c]{$\beta=0.1$}& \makecell[c]{$84.85\pm0.86$}  & \makecell[c]{$84.51\pm0.86$} & \makecell[c]{$79.21\pm0.62$}  & \makecell[c]{$77.94\pm0.80$}&  \makecell[c]{$76.93\pm1.40$}  & \makecell[c]{$75.17\pm2.68$}  & \makecell[c]{$\mathbf{86.90}\pm0.75$} &\makecell[c]{$\mathbf{86.70}\pm0.85$}  \\
 \bottomrule
\end{tabular}
\label{tab:heterogeneity}
\vspace{-2mm}
\end{table*}

\begin{table*}[!t]
\center
\renewcommand\arraystretch{1.0}
\setlength{\tabcolsep}{2pt}
\caption{The performance comparison of different methods with different ranks of LoRA.
}
\begin{tabular}{p{40pt}|p{45pt}|p{45pt}|p{45pt}|p{48pt}|p{48pt}|p{45pt}|p{46pt}|p{48pt}|p{48pt}}
\toprule
\multirow{2}{1.2cm}{\makecell[c]{Datasets} }&\multirow{2}{1.7cm}{\centering\makecell[c]{LoRA Rank} } & \multicolumn{2}{l|}{\makecell[c]{LoRA}} & \multicolumn{2}{l|}{\makecell[c]{FFA-LoRA}}& \multicolumn{2}{l|}{\makecell[c]{DP-DyLoRA}} & \multicolumn{2}{l}{\makecell[c]{DEeR}} \\
\cmidrule(r){3-10}
&                 &    \makecell[c]{Accuracy}     &   \makecell[c]{F1-score }   &   \makecell[c]{Accuracy}        &  \makecell[c]{F1-score} &   \makecell[c]{Accuracy}      &   \makecell[c]{F1-score} &   \makecell[c]{Accuracy}      &   \makecell[c]{F1-score}   \\
\midrule
\multirow{4}{1cm}{\makecell[c]{OCT-8} }&\makecell[c]{ $r=16$}&\makecell[c]{$46.77\pm5.07$}  &  \makecell[c]{$41.53\pm5.74$} &  \makecell[c]{$56.85\pm4.21$} & \makecell[c]{$54.70\pm5.70$}  & \makecell[c]{$30.28\pm4.00$} & \makecell[c]{$22.81\pm2.13$}    &  \makecell[c]{$\mathbf{87.50}\pm1.99$} & \makecell[c]{$\mathbf{87.31}\pm2.16$}  \\
&\makecell[c]{ $r=8$} & \makecell[c]{$42.15\pm2.88$}  &\makecell[c]{$38.59\pm3.60$}  & \makecell[c]{$56.20\pm4.63$} & \makecell[c]{$52.29\pm6.10$} &  \makecell[c]{$23.55\pm4.61$} & \makecell[c]{$19.30\pm3.40$} & \makecell[c]{$\mathbf{84.28}\pm3.74$} & \makecell[c]{$\mathbf{83.20}\pm4.59$}   \\
&\makecell[c]{ $r=4$}& \makecell[c]{$35.34\pm4.38$}  &\makecell[c]{$29.33\pm2.48$} & \makecell[c]{$66.32\pm6.56$} & \makecell[c]{$64.61\pm8.29$}&  \makecell[c]{$21.54\pm5.43$} & \makecell[c]{$16.38\pm6.33$} & \makecell[c]{$\mathbf{78.09}\pm3.26$}  & \makecell[c]{$\mathbf{76.21}\pm4.21$}  \\
&\makecell[c]{ $r=2$}&\makecell[c]{$23.07\pm1.81$}  &\makecell[c]{$17.28\pm0.18$}  & \makecell[c]{$\mathbf{63.55}\pm4.84$} & \makecell[c]{$\mathbf{60.25}\pm5.32$} &  \makecell[c]{$19.95\pm1.84$} & \makecell[c]{$12.80\pm2.11$} & \makecell[c]{$61.28\pm3.69$} & \makecell[c]{$57.11\pm5.35$} \\
\midrule
\multirow{4}{1cm}{\makecell[c]{Kvasir-v2} }&\makecell[c]{ $r=16$}&\makecell[c]{$83.39\pm1.04$}  &\makecell[c]{$82.81\pm1.16$}  & \makecell[c]{$76.75\pm0.68$} & \makecell[c]{$75.50\pm0.95$}&  \makecell[c]{$81.97\pm1.28$}  &  \makecell[c]{$81.15\pm1.34$}  & \makecell[c]{$\mathbf{86.81}\pm0.75$} & \makecell[c]{$\mathbf{86.67}\pm0.78$} \\
&\makecell[c]{ $r=8$} & \makecell[c]{$84.85\pm0.86$}  & \makecell[c]{$84.51\pm0.86$} & \makecell[c]{$79.21\pm0.62$}  & \makecell[c]{$77.94\pm0.80$}& \makecell[c]{$76.93\pm1.40$}  & \makecell[c]{$75.17\pm2.68$} & \makecell[c]{$\mathbf{86.90}\pm0.75$} &\makecell[c]{$\mathbf{86.70}\pm0.85$}  \\
&\makecell[c]{ $r=4$} &\makecell[c]{$84.43\pm0.40$}  & \makecell[c]{$84.14\pm0.36$}&\makecell[c]{$79.56\pm0.38$}  & \makecell[c]{$78.40\pm0.13$}&  \makecell[c]{$75.10\pm4.65$}  & \makecell[c]{$71.37\pm7.37$} & \makecell[c]{$\mathbf{87.42}\pm0.38$} & \makecell[c]{$\mathbf{87.25}\pm0.45$} \\
&\makecell[c]{ $r=2$}&\makecell[c]{$79.35\pm0.77$}  &\makecell[c]{$77.97\pm1.38$}  & \makecell[c]{$79.68\pm0.70$} & \makecell[c]{$78.58\pm0.70$} & \makecell[c]{$72.16\pm2.27$}  &\makecell[c]{$69.26\pm3.44$}& \makecell[c]{$\mathbf{84.83}\pm1.15$} & \makecell[c]{$\mathbf{84.61}\pm1.27$} \\
 \bottomrule
\end{tabular}
\label{tab:rank}
\vspace{-3mm}
\end{table*}

\begin{table*}[!t]
\center
\renewcommand\arraystretch{1.0}
\setlength{\tabcolsep}{2.5pt}
\caption{The performance comparison of different methods with different client number. 
}
\begin{tabular}{p{40pt}|p{45pt}|p{45pt}|p{45pt}|p{45pt}|p{45pt}|p{45pt}|p{45pt}|p{45pt}|p{45pt}}
\toprule
\multirow{2}{1.2cm}{\makecell[c]{Datasets} }&\multirow{2}{1.7cm}{\centering\makecell[c]{LoRA Rank} } & \multicolumn{2}{l|}{\makecell[c]{LoRA}} & \multicolumn{2}{l|}{\makecell[c]{FFA-LoRA}}& \multicolumn{2}{l|}{\makecell[c]{DP-DyLoRA}} & \multicolumn{2}{l}{\makecell[c]{DEeR}} \\
\cmidrule(r){3-10}
                  &                   & \makecell[c]{Accuracy} &  \makecell[c]{F1-score}  &   \makecell[c]{Accuracy} &  \makecell[c]{F1-score}       &  \makecell[c]{Accuracy} &  \makecell[c]{F1-score}   &    \makecell[c]{Accuracy} &  \makecell[c]{F1-score}        \\
\midrule
\multirow{4}{1cm}{\makecell[c]{OCT-8} }
                  &\makecell[c]{ $K=16$}&\makecell[c]{$52.71\pm3.13$} &\makecell[c]{$49.48\pm3.39$} &\makecell[c]{$64.01\pm2.51$} & \makecell[c]{$62.52\pm2.65$}  & \makecell[c]{$32.60\pm4.72$} & \makecell[c]{$27.39\pm5.72$} &\makecell[c]{$\textbf{88.57}\pm1.34$} &\makecell[c]{$\textbf{88.34}\pm1.46$} \\
                  &\makecell[c]{ $K=12$}&\makecell[c]{$42.15\pm2.88$} &\makecell[c]{$38.59\pm3.60$} &\makecell[c]{$56.20\pm4.63$} & \makecell[c]{$52.29\pm6.10$}  & \makecell[c]{$23.55\pm4.61$} &\makecell[c]{$19.30\pm3.40$} &\makecell[c]{$\textbf{84.28}\pm3.74$} &\makecell[c]{$\textbf{83.20}\pm4.59$} \\
                  &\makecell[c]{ $K=8$}&\makecell[c]{$30.57\pm3.84$} &\makecell[c]{$21.94\pm3.35$} &\makecell[c]{$54.42\pm2.17$} & \makecell[c]{$49.91\pm2.04$}  & \makecell[c]{$22.08\pm3.96$} &\makecell[c]{$16.58\pm2.51$} &\makecell[c]{$\textbf{88.13}\pm0.79$} &\makecell[c]{$\textbf{87.89}\pm0.82$} \\
                  &\makecell[c]{ $K=4$}&\makecell[c]{$25.91\pm1.03$} &\makecell[c]{$19.77\pm2.31$} &\makecell[c]{$42.16\pm6.02$} & \makecell[c]{$34.99\pm7.88$}  & \makecell[c]{$22.30\pm1.21$} &\makecell[c]{$15.32\pm2.67$} &\makecell[c]{$\textbf{76.39}\pm4.74$} &\makecell[c]{$\textbf{73.39}\pm6.41$} \\
\midrule
\multirow{4}{1cm}{\makecell[c]{Kvasir-v2}}
                  &\makecell[c]{ $K=16$}&\makecell[c]{$85.58\pm0.26$} &\makecell[c]{$85.46\pm0.34$} &\makecell[c]{$82.27\pm1.88$} & \makecell[c]{$82.15\pm1.89$}  & \makecell[c]{$84.75\pm1.45$} &\makecell[c]{$84.50\pm1.58$} &\makecell[c]{$\textbf{87.70}\pm0.69$} &\makecell[c]{$\textbf{87.61}\pm0.66$} \\
                  &\makecell[c]{ $K=12$}&\makecell[c]{$84.85\pm0.86$} &\makecell[c]{$84.51\pm0.86$} &\makecell[c]{$79.21\pm0.62$} & \makecell[c]{$77.94\pm0.80$}  & \makecell[c]{$76.93\pm1.40$} &\makecell[c]{$75.17\pm2.68$} &\makecell[c]{$\textbf{86.90}\pm0.75$} &\makecell[c]{$\textbf{86.70}\pm0.85$} \\
                  &\makecell[c]{ $K=8$}&\makecell[c]{$81.75\pm1.48$} &\makecell[c]{$80.81\pm2.02$} &\makecell[c]{$79.64\pm1.62$} & \makecell[c]{$78.83\pm2.42$}  & \makecell[c]{$75.81\pm0.40$} &\makecell[c]{$73.67\pm1.20$} &\makecell[c]{$\textbf{86.95}\pm0.83$} &\makecell[c]{$\textbf{86.79}\pm0.91$} \\
                  &\makecell[c]{ $K=4$}&\makecell[c]{$69.72\pm1.44$} &\makecell[c]{$64.71\pm1.13$} &\makecell[c]{$68.66\pm1.45$} & \makecell[c]{$64.37\pm1.61$}  & \makecell[c]{$59.66\pm2.90$} &\makecell[c]{$52.83\pm4.28$} &\makecell[c]{$\textbf{78.47}\pm2.19$} &\makecell[c]{$\textbf{75.16}\pm2.93$} \\
 \bottomrule
\end{tabular}
\label{tab:client_number}
\vspace{-3mm}
\end{table*}

\subsection{Ablation Analysis}
We perform a comprehensive evaluation on Kvasir-v2 and OCT-8 to investigate the efficacy of different modules in DEeR and the impact of some critical factors, i.e., data heterogeneity $\beta$, rank $r$, communication budget and client number.
\subsubsection{Evaluation of Different Modules}
The deviation eliminator and noise regulator are two indispensable components for DEeR to improve the finetuning performance. To evaluate their contributions, we individually remove them to observe the performance of DEeR. As illustrated  in Table~\ref{tab:ablation}, DEeR experiences significant performance decline once we remove deviation eliminator (w/o Deviation Eliminator), with decrements of $48.5\%$ from $92.33\%$ to $43.83\%$ on OCT-8 ($\varepsilon = 1.0$)  and $14.57\%$ from $87.00\%$ to $72.43\%$ on Kvasir-v2 ($\varepsilon = 6.0$) in accuracy. The decrements are further magnified to $68.39\%$ and $20.63\%$ when the budgets $\varepsilon$ on two datasets shrink to $0.1$ and $1.0$, respectively.  Moreover, we can observe that discarding noise regulator (w/o Noise Regulator) leads to a slight performance drop when the privacy budget is high on Kvasir-v2 ($\varepsilon = 6.0$) and OCT-8 ($\varepsilon = 1.0$). Nonetheless, severe performance degradation is triggered by a lower budget, especially on Kvasir-v2. The best results are obtained when DEeR is equipped with deviation eliminator and noise regulator, which can corroborate the effectiveness of the two modules.

\subsubsection{Impact of Data Heterogeneity}
Based on the analysis of Theorem 1, the data heterogeneity can exacerbate aggregation deviation. To investigate its effect, we use the default privacy budget $\varepsilon$ and change the heterogeneity parameter $\beta$ to observe the performance of different methods in Table \ref{tab:heterogeneity}. The results show that LoRA and FFA-LoRA undergo more considerable performance drop with decreasing $\beta$ compared with DEeR. For example, on OCT-8, as $\beta$ decreases from $10.0$ to $0.1$, F1-scores of LoRA and FFA-LoRA drop from $70.69\%$ to $38.59\%$ with a decrement of $32.10\%$, and $86.19\%$ to $52.29\%$ with a decrement of $33.90\%$, respectively. Noticeably, the decrement of F1-score for DEeR is merely $11.66\%$. We can find significant difference in DEeR and the second-best FFA-LoRA with $P$-value $<0.005$ for all $\beta$. Besides, DEeR achieves better performance than LoRA, FFA-LoRA and DP-DyLoRA on two datasets for different $\beta$. The performance advantage can prove that DEeR is more robust against data heterogeneity than existing methods and further confirms the effectiveness of the proposed noise regulator.
\subsubsection{Impact of Rank $\textit{r}$}
The rank $r$ can be regarded as LoRA parameter budget. A larger $r$ indicates more trainable parameters. We fix the privacy budget $\varepsilon$ as well as the heterogeneity parameter $\beta$, and compare the performance of different methods with various $r \in [2, 4, 8, 16]$. As presented in Table~\ref{tab:rank}, FFA-LoRA achieves the best performance with $63.55\%$ in accuracy and $60.25\%$ in F1-score on OCT-8 when $r=2$. As $r$ increases from 2 to 16, DEeR exhibits promising performance improvement to $87.50\%$ and $87.31\%$ with increments of $26.22\%$ and $30.20\%$ in accuracy and F1-score respectively, while accuracy and F1-score of FFA-LoRA decrease to $56.85\%$ and $54.70\%$ ($P$-value $<0.005$ for $r=8$ and $P$-value $<0.002$ for $r=16$), and DP-DyLoRA always present extremely poor performance.
For Kvasir-v2, all methods do not show huge performance fluctuations as $r$ changes. It is worth mentioning that DEeR exceeds LoRA and FFA-LoRA by a huge margin for different $r$. Particularly, DEeR with the fewest parameters ($r=2$) can obtain better performance than the second-best method (LoRA) with the most parameters ($r=16$). Significant difference is presented in DEeR and the second-best LoRA for $r=2$ ($P$-value $<0.01$) and $r=4$ ($P$-value $<0.005$). DP-DyLoRA shows not only bigger performance fluctuations as $r$ changes but also lower performance than DEeR.

\subsubsection{Impact of Communication Budget}
\begin{figure}[!h]	
	\begin{center}
		\subfigure[Variants of gAM optimization algorithm]
		{\includegraphics[width=1.0\columnwidth]{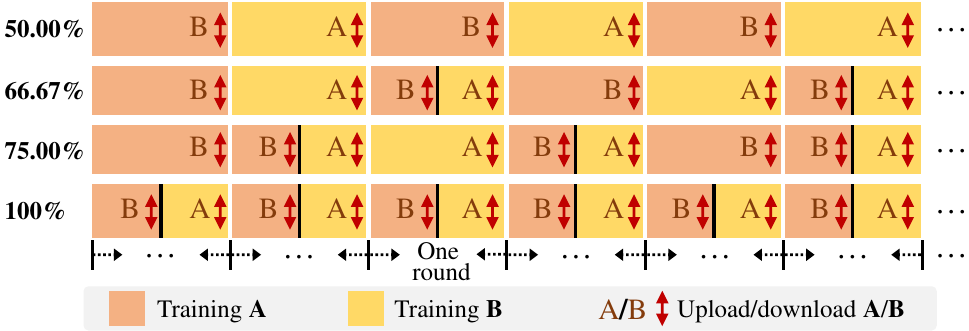}\label{fig:gAM}}
		\subfigure[OCT-8]
		{\includegraphics[width=0.485\columnwidth]{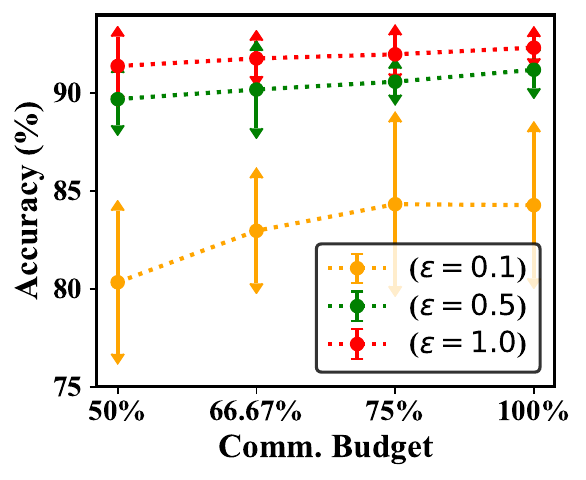}\label{fig:oct_acc}}
		\subfigure[Kvasir-v2]
		{\includegraphics[width=0.495\columnwidth]{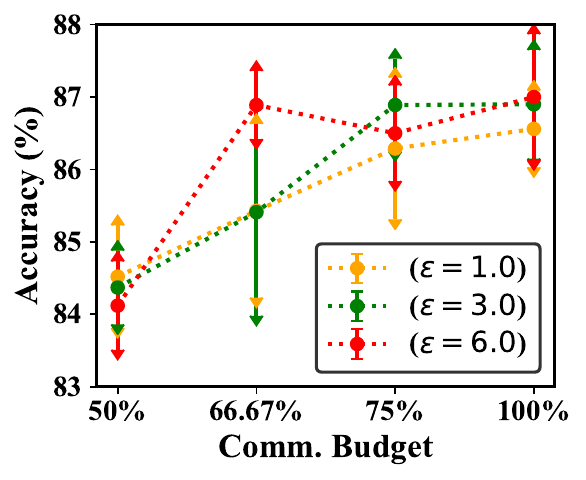}\label{fig:kav_acc}}
	\end{center}
	\vspace{-2mm}
	\caption{The impact of communication budgets in different variants of gAM optimization algorithm.}\label{fig:comm_budget}
	\vspace{-4mm}
\end{figure}

In DEeR, different variants of gAM optimization algorithm lead to different communication budgets, as shown in Fig. \ref{fig:gAM}. We assume that the communication budget is $100\%$ when one round of training includes both $\mathbf{A}$ and $\mathbf{B}$, while it is $50\%$ if one round only contains $\mathbf{A}$ or $\mathbf{B}$. Fig.~\ref{fig:comm_budget}(b)-(c) present performance of different variants on OCT-8 and Kvasir-v2, respectively. In Fig.~\ref{fig:comm_budget}(b), even though we reduce the communication budget from $100\%$ to $50\%$, model still maintains a good and stable performance for big privacy budgets ($\varepsilon = 0.5$ or $1.0$) on OCT-8 dataset. Fig.~\ref{fig:comm_budget}(c) demonstrates that $75\%$ of the communication budget can achieve similar performance as $100\%$ of the budget for all privacy budgets on Kvasir-v2 dataset. Moreover, the best performance from $100\%$ of the communication budget indicates the importance of optimizing both $\mathbf{A}$ and $\mathbf{B}$ for each round.

\subsubsection{Impact of Client Number} To compare the performance of different methods across different numbers of clients, we fix the privacy budgets $\varepsilon$, the data heterogeneity $\beta$ and the rank $r$ of LoRA layers, and divide the training data of OCT-8 and Kvasir-v2 datasets into $K$ clients, respectively. As shown in Table~\ref{tab:client_number}, for OCT-8 dataset, all existing methods yield limited classification performance for various client numbers. DEeR exceeds the second-best method FFA-LoRA by large margins, such as 34.23\% ($P$-value $<0.002$) and 24.56\% ($P$-value $<0.05$) in Accuracy for $K$ = 4 and 16, respectively. Meanwhile, DEeR also significantly outperforms these methods for any $K$ on Kvasir-v2 dataset. For both two datasets, DEeR shows a lower sensitivity to the client number $K$ than previous methods. The performance gap of these methods between $K$=4 and 16 surpasses 20\% in terms of  Accuracy and F1-score, while the gap of DEeR is around 10\%. Therefore, these results illustrate that DEeR is more robust against the number of clients than existing approaches.
\section{Conclusion}
\label{sec:conclusion}
In this paper, we propose a novel FedFT framework named DEeR, which exploits LoRA to adapt pretrained foundation models to downstream medical tasks in FL with client-level DP guarantees. We first delve into two challenges of FedFT with LoRA, \textit{i.e.}, aggregation deviation and noise amplification effect. Afterwards, a deviation eliminator is proposed to utilize the alternating minimization optimization algorithm to iteratively optimize the parameters of LoRA for avoiding aggregation deviation. Besides, we present a noise regulator at the client side that introduces two regulator factors  to suppress the noise amplification effect. The comprehensive experiments on two classification and two segmentation datasets validate the effectiveness of DEeR. The results show DEeR achieves superior performance than state-of-the-art methods. The ablated experiments verify the importance of key modules in DEeR, investigate the impact of data heterogeneity, rank $r$, the communication budget and client number.

The proposed DEeR has achieved promising performance on various medical tasks, yet there are several limitations: (1) In DEeR, gAM algorithm is exploited to optimize the parameters of LoRA for avoiding aggregation deviation. However, the alternating optimization strategy will increase the training time. Although we have explored different variants of gAM algorithm to reduce communication frequency, they fail to achieve the same performance as 100\% of the communication frequency for different privacy budgets. (2) DEeR may undergo data security risk during the communication process. Although our method does not share the raw data of clients and protects client-level privacy by DP, local client models might be stolen by intruders and competitors for the reconstruction of training data~\cite{kairouz2021advances}. For this problem, we are able to apply existing homomorphic encryption techniques~\cite{kairouz2021advances} to encrypt client models and the global model.

\bibliographystyle{IEEEtran}
\bibliography{bare_jrnl}



%

\end{document}